\documentclass[10pt]{article} 
\usepackage[accepted]{tmlr}

\usepackage{amsmath,amsfonts,bm}

\def\eqref#1{equation~\ref{#1}}

\def\1{\bm{1}}

\DeclareMathAlphabet{\mathsfit}{\encodingdefault}{\sfdefault}{m}{sl}
\SetMathAlphabet{\mathsfit}{bold}{\encodingdefault}{\sfdefault}{bx}{n}

\usepackage{hyperref}
\usepackage{url}
\usepackage{capt-of}
\usepackage[utf8]{inputenc} 
\usepackage[T1]{fontenc}    
\usepackage{booktabs}       
\usepackage{amsfonts}       
\usepackage{tabularx} 
\usepackage{nicefrac}       
\usepackage{microtype}      
\usepackage{amsmath}
\usepackage{amssymb}
\usepackage{multirow}
\usepackage{subcaption}
\usepackage{pifont}
\usepackage{algorithm}
\usepackage{algpseudocode}
\usepackage{bbm}
\usepackage{textcomp}
\usepackage{stfloats}
\usepackage{mathabx}
\usepackage{amsthm}
\usepackage{enumitem}
\setlist{leftmargin=5mm}
\usepackage{wrapfig}
\usepackage{makecell}
\usepackage{graphicx}
\usepackage{xcolor, xspace}
\usepackage{tcolorbox}
\usepackage{float}
\usepackage[table]{xcolor}
\newtcolorbox{mybox}[1][]{
    title=#1,
    fonttitle=\small,
    fontupper=\small,
    left=2mm,
    right=2mm,
    top=1mm,
    bottom=0mm,
}
\usepackage{inconsolata}

\usepackage{tikz}
\makeatletter
\newcommand{\DrawLine}{%
  \begin{tikzpicture}
  \path[use as bounding box] (0,0) -- (\linewidth,0);
  \draw[color=black,dashed,dash phase=2pt]
        (0-\kvtcb@leftlower-\kvtcb@boxsep,0)--
        (\linewidth+\kvtcb@rightlower+\kvtcb@boxsep,0);
  \end{tikzpicture}%
  }

\newcommand{\C}{\mathcal{C}}
\newcommand{\Ibase}{\mathcal{I}_{\text{base}}}
\newcommand{\Iprime}{\mathcal{I}'}

\def\huggingface{\raisebox{-1.5pt}{\includegraphics[height=1.05em]{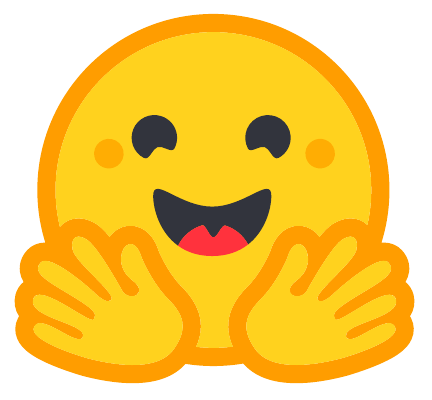}}}
\def\github{\raisebox{-1.5pt}{\includegraphics[height=1.0em]{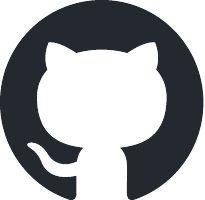}}}
  
\definecolor{violet}{RGB}{138, 43, 226}

\makeatother

\definecolor{citepcol}{HTML}{2DDC0E}
\definecolor{tableofcontent}{HTML}{E63E15}
\definecolor{urlcol}{HTML}{2470D8}
\definecolor{myorange}{RGB}{2, 142, 2}
\hypersetup{
    colorlinks=true,
    linkcolor=red,
    filecolor=magenta,      
    urlcolor=cyan,
}

\newcommand\ours{\textsc{TS-Reasoner}\xspace}
\title{TS-Reasoner: Aligning Time Series Foundation Models with LLM Reasoning}

\author{\name Fangxu Yu \email yufx@umd.edu \\
      \addr University of Maryland, College Park
      \AND
      \name Hongyu Zhao \email hongyuz@umd.edu \\
      \addr University of Maryland, College Park
      \AND
      \name Tianyi Zhou \email Tianyi.Zhou@mbzuai.ac.ae\\
      \addr Mohamed bin Zayed University of Artificial Intelligence  
}

\def\month{08}  
\def\year{2026} 
\def\openreview{\url{https://openreview.net/forum?id=d6TD0f2xXq}} 

\begin{document}

\maketitle

\vspace{-20pt}
\makebox[\linewidth][c]{%
\huggingface\;\href{https://huggingface.co/ParadiseYu/TS-Reasoner-7B}{Model Weights}\hspace{2em}
\huggingface\;\href{https://huggingface.co/datasets/ParadiseYu/TS-Reasoner-trainingset}{Datasets}\hspace{2em}
\github\;\href{https://github.com/Yu-Fangxu/TS-Reasoner}{Code}}

\begin{abstract}
Time series reasoning is crucial to decision-making in diverse domains, including finance, energy, and scientific discovery. While existing time series foundation models (TSFMs) can capture low-level dynamic patterns and provide accurate forecasting, further analysis usually requires additional background knowledge and sophisticated reasoning, which are lacking in most TSFMs but can be achieved through Large Language Models (LLMs). On the other hand, without expensive post-training, LLMs often struggle with the numerical understanding of time series data. Although it is intuitive to integrate the two types of models, developing effective training recipes that align the two modalities for reasoning tasks is still an open challenge.
To this end, we propose \ours that aligns the latent representations of TSFMs with the textual inputs of LLMs for downstream understanding/reasoning tasks. 
Specifically, we propose a simple yet effective method to curate diverse, synthetic pairs of time series and textual captions for alignment training. We then develop a two-stage training recipe that applies instruction fine-tuning after the alignment pretraining. Unlike existing works that train an LLM to take time series as inputs, we leverage a pretrained TSFM and freeze it during training. 
Experiments on several benchmarks demonstrate that \ours not only outperforms a wide range of open-source LLMs, Vision Language Models (VLMs), and Time Series LLMs of comparable scale, but also achieves this with remarkable data efficiency, e.g., using less than half the training data.

\end{abstract}
\section{Introduction}
Time series analysis has long been fundamental to various real-world applications in finance, energy, weather, traffic, and other domains~\citep{prakarsha2022time, xu2023density, nie2024survey}. Its ability to model dynamics and predict future states based on historical data makes it an indispensable tool for decision-making and strategic planning. While numerical attributes form the bedrock of time series analysis, human decision-making is often complemented by rich prior knowledge and qualitative contextual information, including news articles, social media trends, and expert assessments. This gap prevents analytical models from achieving a deeper, more contextualized understanding of the events and dynamics driving the numerical data. By enabling machines to understand both contextual information and numerical time series patterns, we can empower them as automated systems that assist humans in gaining deeper insights into complex phenomena.

\begin{figure*}[t]   
    \centering
    \includegraphics[width=0.7\textwidth]{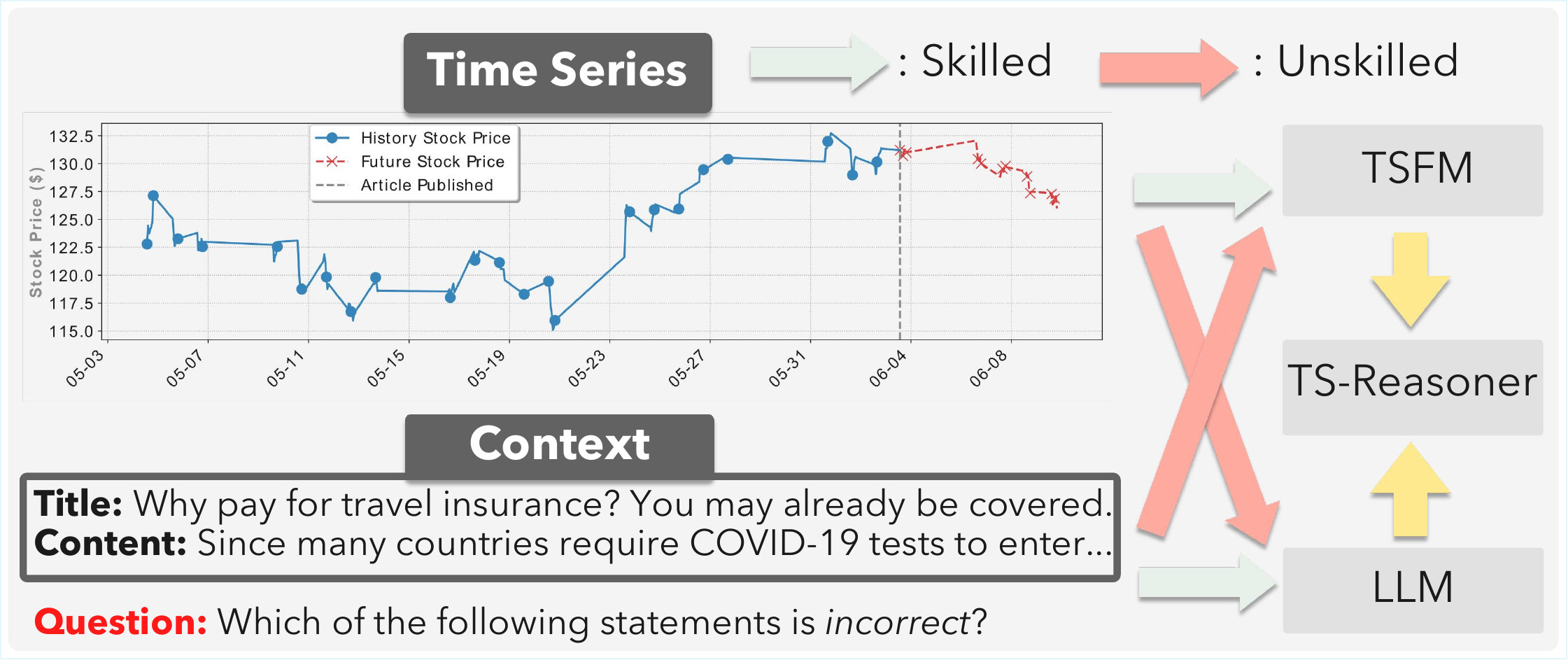}
    \vspace{-5pt}
    \caption{Time series forecasting vs. reasoning. The time series reasoning task requires both contextual reasoning (e.g., news) by LLMs and numerical understanding by TSFM.}
    \label{fig:task_intro}
    \vspace{-20pt}
\end{figure*}

Recent advances in Time Series Foundation Models (TSFMs) have significantly enhanced the understanding of time series data through large-scale pretraining. These models are capable of generalizing across various time series tasks and domains. Although TSFMs~\citep{goswami2024moment, das2024decoder} demonstrate strong modeling capabilities, most are pre-trained exclusively on unimodal numerical time series and cannot therefore comprehend or integrate textual information. On the other hand, Large Language Models (LLMs) and Vision Language Models (VLMs) can take texts and images as input context, and have demonstrated remarkable reasoning and problem-solving abilities across various tasks~\citep{wei2022chain, hao2023reasoning, yu2024flow, ho2025arcmemo, yu2026weak}, sparking interest in transferring their capabilities to time series analysis. Some studies~\citep{gruver2023large, liu2024lstprompt, jia2024gpt4mts} transform numerical time series into string form and perform time series forecasting on LLMs by prompting them with the strings. However, despite their strong reasoning abilities, LLMs struggle to capture temporal dependencies due to their inherent lack of temporal understanding~\citep{fons2024evaluating, merrill2024language} and limited ability to interpret numerical values. These limitations hinder their understanding of time series data. As shown in Figure~\ref{fig:task_intro}, TSFM and LLM have complementary strengths; the former specializes in temporal understanding, while the latter excels at text understanding. 
To combine the complementary strengths of TSFM and LLM while overcoming their respective limitations, we propose \ours, a Time Series Large Language Model (TSLLM) designed to enhance time series reasoning by aligning a TSFM with an LLM. Specifically, we first employ the TSFM to extract rich temporal representations from numerical time series data. To effectively incorporate this temporal information into the LLM, \ours introduces a TS-to-Text adapter, which projects the TSFM-extracted temporal features into the LLM’s input embedding space. This enables seamless integration of the TSFM's temporal understanding with the LLM’s powerful linguistic and reasoning capabilities. Our training framework consists of two stages: pretraining and fine-tuning. In the pretraining stage, we finetune \ours to produce textual captions of input time series and achieve a fundamental alignment.
To this end, we propose a simple yet effective prompting strategy to curate high-quality captions for diverse time series data using advanced LLMs/VLMs. 
In the fine-tuning stage, we further enhance the model’s reasoning abilities through instruction tuning, ensuring robust performance in downstream tasks.

\begin{wrapfigure}{r}{0.6\textwidth}
\vspace{-20pt}
    \centering
    \includegraphics[width=0.6\textwidth]{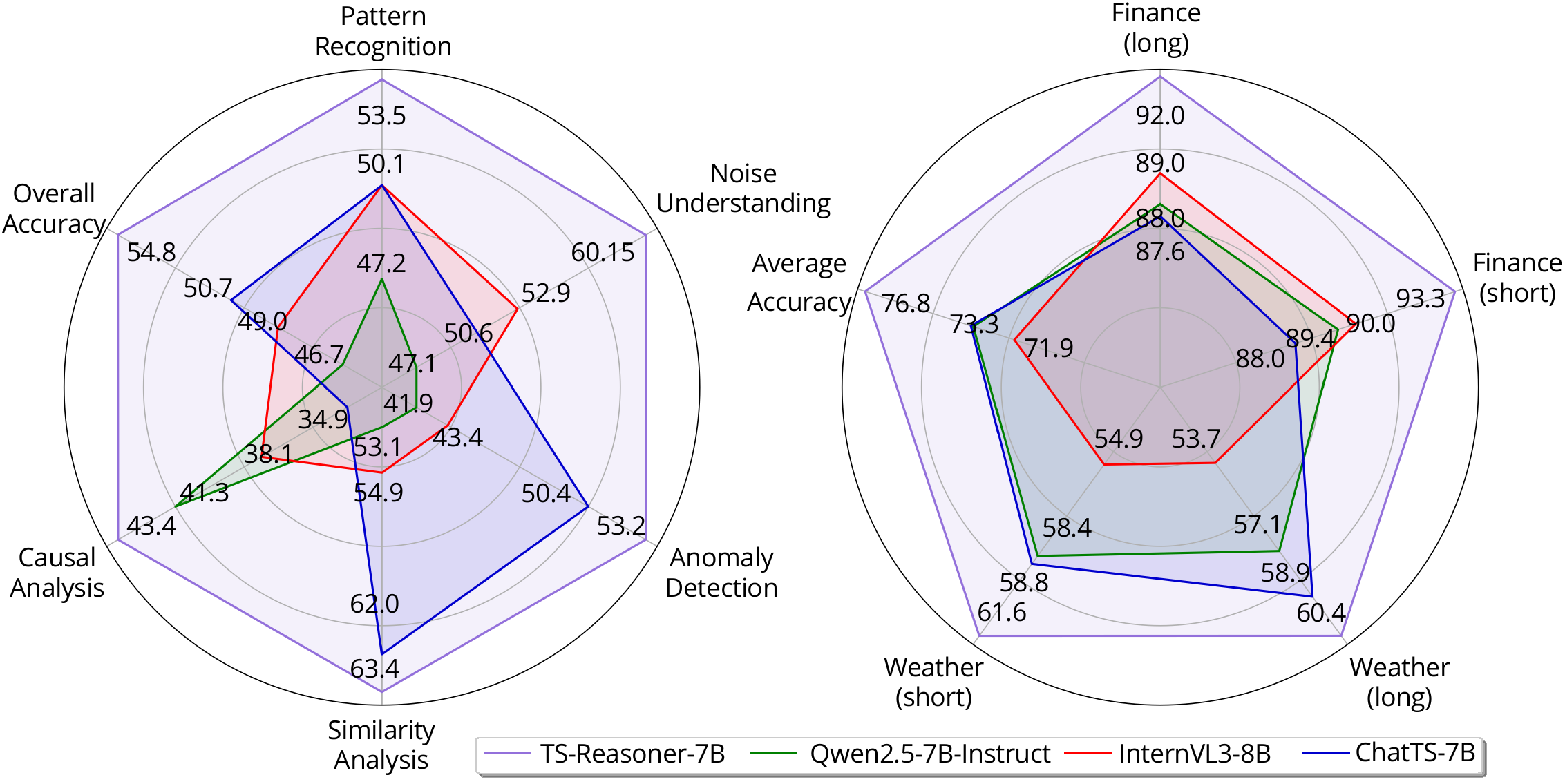}
    \caption{Results on time series understanding and reasoning benchmarks. \ours shows a consistent advantage over the comparable-scale open-source LLMs, VLMs, and TSLLMs.}
    \label{fig:radar}
    \vspace{-15pt}
\end{wrapfigure}
Our work makes unique contributions to a recent line of research combining TSFMs and LLMs. First, our formulation sets up the connection between LLMs and TSFMs, facilitating time series reasoning through the integration of rich contextual information and LLM reasoning. 
Second, we address a critical data bottleneck with a simple yet effective time series captioning method, which diversifies the training data for aligning LLMs and TSFMs. Finally, we offer new empirical insights into the strengths and limitations of existing approaches.

We evaluate the understanding and reasoning capabilities of our approach on two standard benchmarks: TimeSeriesExam~\citep{cai2024timeseriesexam} and MTBench~\citep{chen2025mtbench}. \ours significantly outperforms a wide range of open-source models, including LLMs, VLMs, and the TSLLMs, as shown in Figure~\ref{fig:radar}. Finally, comprehensive analyses, including extensive ablation studies, validate the effectiveness of our key designs and establish the superiority of \ours in generalization performance, scalability, and training data efficiency.

\section{Related Work}
\textbf{LLMs for Time Series.} 
LLMs have recently garnered significant interest in time series analysis. Traditional time series forecasting relies on statistical models~\citep{rb1990stl} or data-driven neural networks~\citep{liu2021pyraformer, lim2021temporal, wu2021autoformer, zhou2022fedformer, li2023prototype, li2024transformer} for tasks like weather and stock prediction. Recent efforts explore LLMs for this task, with some designing prompts to elicit forecasting abilities~\citep{cao2023tempo, ltsm-bundle}. Others focus on enabling LLMs to understand time series data by converting it into textual sequences or aligning its embeddings with language model embeddings via prompting or semantic information~\citep{jin2023time, sun2023test, pmlr-v235-pan24c}. In addition, multimodal vision-based LLMs are being investigated for time series prediction~\citep{chen2024visionts, zhong2025time, yu2026tsrbench, yu2026tsrouter}.
Though LLMs exhibit non-trivial performance on some forecasting tasks, \citep{merrill2024language} indicates that LLMs struggle to reason about time series. To tackle this challenge, several works~\citep{chow2024towards, zhang2025tempogpt, xie2024chatts} propose to enable LLMs to understand the time series with context. \ours lies in this direction, distinguishing itself by employing a pre-trained Time Series Foundation Model to ground the LLM's reasoning in robust temporal features. See Appendix~\ref{sec:arch-comparison} for a comparison with representative architectures that leverage LLMs/VLMs for time series.

\textbf{Modality Alignment.} Modality alignment methods are widely studied in the multimodal domain~\citep{li2022blip, lai2024veclip, li2023blip, liu2024visual, yu2026arrowgev, yu2026reinforcement}. Inspired by the success of multimodal alignment, recent works treat time series as another modality and align it to the LLM~\citep{xie2024chatts, zhang2025tempogpt}. Though they achieve a certain degree of time series understanding, they focus on narrow domains (e.g., electricity) and tasks (e.g., time series understanding), and train time series encoders from scratch. In contrast, we adapt the successful training paradigm in VLMs, identify and address the key challenges (e.g., integration of characteristics of time series into LLMs, and the shortage of time series-text pairs) faced in applying this paradigm to the unique modality of time series, exploring pre-trained time series foundation models to exploit rich time series knowledge.

\textbf{Time Series Foundation Models.}
Recent advancements in pre-training methods are significantly contributing to the development of foundation models for time series analysis. Early efforts, such as TST~\citep{zerveas2021transformer} and PatchTST~\citep{nie2022time}, applied BERT-like masked pretraining techniques, focusing on point-level and patch-level masking, respectively. A separate line of work, exemplified by models like TimesFM~\citep{das2024decoder}, Timer~\citep{liu2024timer}, TTMS~\citep{ekambaram2024tiny}, Chronos~\citep{ansari2024chronos}, Time-MoE~\citep{shi2024time}, Moirai~\citep{liu2024moirai}, CoRA~\citep{qin2025cora}, TimesBERT~\citep{zhang2025timesbert}, and Sundial~\citep{liu2025sundial} demonstrates the advantages of large-scale pre-training for improving forecasting performance. Exploring diverse pre-training objectives, MOMENT~\citep{goswami2024moment} leverages a T5 encoder to achieve strong downstream multi-task capabilities. ChronoSteer~\citep{wang2025chronosteer} also explores the alignment between TSFMs and LLMs, yet it leverages the LLM's revisions to enhance TSFM's forecasting capability.
\section{\ours for Temporal Reasoning}

\subsection{Problem Formulation}
We start by defining time series reasoning. Given a natural language context $\mathcal{X}$, which may encode background domain knowledge or prompt instructions, and a corresponding set of time series $\mathcal{S}=\{\mathcal{T}_{1},\dots,\mathcal{T}_{K}\}$, the model produces an output sequence $\mathcal{V}$. To explicitly model the reasoning capability, we formulate $\mathcal{V}$ as a composition of a reasoning path $\mathcal{R}$ and a final answer $\mathcal{A}$, denoted as $\mathcal{V} = [\mathcal{R}; \mathcal{A}]$. Here, $\mathcal{R}$ consists of tokens providing intermediate explanations for problem-solving, and $\mathcal{A}$ contains the tokens for the final conclusion.

The model defines a probability distribution over the output sequence $\mathcal{V}$, conditioned on inputs $\mathcal{X}$ and $\mathcal{S}$. This generation process can be factorized as:
\begin{equation}
    P(\mathcal{V} | \mathcal{X}, \mathcal{S}) = \underbrace{P(\mathcal{A} | \mathcal{R}, \mathcal{X}, \mathcal{S})}_{\text{Answer Generation}} \cdot \underbrace{P(\mathcal{R} | \mathcal{X}, \mathcal{S})}_{\text{Reasoning Process}}
\end{equation}
where $P(\mathcal{R} | \mathcal{X}, \mathcal{S})$ indicates the generation of the reasoning path that explains the logical derivation based on the multimodal context, and $P(\mathcal{A} | \mathcal{R}, \mathcal{X}, \mathcal{S})$ represents the generation of the final answer conditioned on the generated reasoning path.
\subsection{Overview}
\begin{figure*}[t]   
\centering
\includegraphics[width=1.0\textwidth]{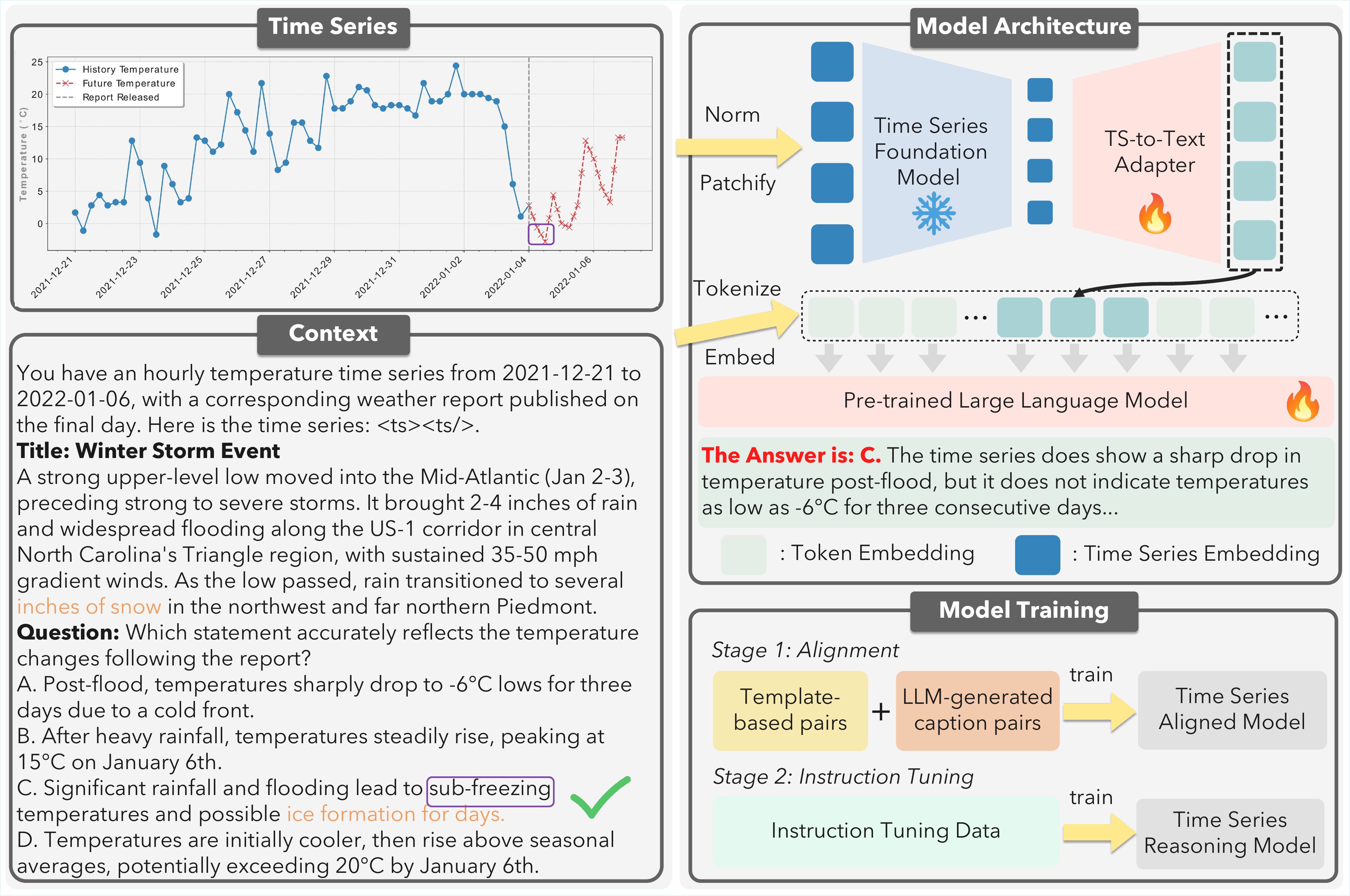}
\caption{Overview of \ours architecture and training pipeline. To perform reasoning, a time series is first encoded by a pretrained Time Series Foundation Model (TSFM). Its output features are then projected into the LLM's input embedding space by a trainable TS-to-Text Adapter and subsequently processed by the LLM. The model is trained in two stages: (i) a pretraining stage that aligns the TSFM outputs with the LLM inputs using both template-based (code-synthesized) and LLM-generated captions, as described in \S\ref{sec: training}, and (2) an instruction-tuning stage to improve complex reasoning capabilities.}
\label{fig:main_arch}
\vspace{-10pt}
\end{figure*}

As illustrated in Figure~\ref{fig:main_arch}, \ours is composed of (i) a pretrained TSFM that encodes normalized, non-overlapping patches of input time series into compact embeddings; (ii) a pretrained LLM, and (iii) a TS-to-Text adapter that projects the TSFM's output embedding to the input space of the LLM. The LLM concatenates the sequence of projected time series features with the sequence of embeddings for input text tokens, with the former demarcated by special tokens ``$\langle\text{ts}\rangle\langle\text{ts}/\rangle$''. 
The training of \ours consists of two stages: (i) a pretraining stage to align time series features from the TSFM with the LLM, using time series caption data synthesized by an advanced LLM/VLM, and (ii) an instruction tuning stage to enhance complex reasoning capabilities on downstream tasks. 

\subsection{Model Architecture}
Given a natural language context $\mathcal{X}$ and a corresponding set of time series 
$\mathcal{S} = \{ \mathcal{T}_1, \mathcal{T}_2, \ldots, \mathcal{T}_{K} \}$, 
we first project both into a shared embedding space. 
Specifically, for each time series $\mathcal{T}_i \in \mathbb{R}^{L_i}$, 
where $L_i$ is the length of the series, we first apply a value-preserved normalization: 
we subtract the mean of the series $\mu_i$, and further divide the centered series by a 
scale factor per series $s_i$ when its magnitude exceeds a fixed stability range. 
This preprocessing step ensures that the model is robust to shifts and scales in the 
input data. Meanwhile, to preserve the absolute magnitude information, the offset $\mu_i$ and 
scale $s_i$ are inserted into the corresponding time-series placeholder in $\mathcal{X}$ 
as plain-text numerical tokens, from which the original values can be exactly recovered. 
Subsequently, we partition the normalized time series into a sequence of non-overlapping 
patches, each of a fixed length $P$. This patching strategy yields a sequence of 
$N_i = \lfloor L_i / P \rfloor$ patches, transforming the time series into a tensor 
$\mathcal{T}_{i}^{p} \in \mathbb{R}^{N_i \times P}$. 
These patches are then encoded using the TSFM, which acts as our time series feature 
extractor. The TSFM processes the sequence of patches and produces a sequence of 
embedding vectors:
\begin{equation}
    \mathcal{Z}_{i}^{T} = \text{TSFM}(\mathcal{T}_{i}^p) \in \mathbb{R}^{N_i \times d_{\text{ts}}},
\end{equation}
where $d_{\text{ts}}$ denotes the dimension of the time series embeddings. Concurrently, the natural language context $\mathcal{X}$ is tokenized and fed into the pre-trained LLM's embedding layer. This process converts the textual input into a sequence of contextualized token embeddings:
\begin{equation}
    \mathcal{Z}^{L} = \text{LLM}_{\text{embed}}(\mathcal{X}) \in \mathbb{R}^{M \times d_{\text{text}}},
\end{equation}
where $M$ is the number of tokens in the instruction, and $d_{\text{text}}$ is the dimensionality of the LLM's hidden states. To align the dimension and semantics of embeddings between LLM and TSFM, we use a single-layer multilayer perceptron (MLP) as a TS-to-Text Adapter to transform the time series embedding into the text embedding space:
\begin{equation}
     \mathcal{H}_{i}^{T} = \text{MLP}(\mathcal{Z}_{i}^T) \in \mathbb{R}^{N_i \times d_{\text{text}}},
\end{equation}
To form a unified input sequence for the LLM that accommodates multiple time series, we structure the natural language instruction $\mathcal{X}$ to include $K$ indicators, $\{ K \cdot \langle \text{ts} \rangle \langle \text{ts/} \rangle \}$. The $i$-th placeholder $\langle\text{ts}\rangle\langle\text{ts/}\rangle$ marks the insertion point for the corresponding $i$-th time series $\mathcal{T}_i$. 

Let $\{\mathcal{H}_i^T \in \mathbb{R}^{N_i \times d_{\text{text}}}\}_{i=1}^K$ be the set of projected time series embeddings, The final input sequence $\mathcal{H}$ is constructed by sequentially inserting the embedding to each  $\langle\text{ts}\rangle\langle\text{ts/}\rangle$ with its corresponding time series embedding sequence $\mathcal{H}_i^T$. 
This substitution process results in a composite sequence where language and time series representations are interleaved. The total length of this fused sequence is $M + \sum_{i=1}^K N_i$. The final tensor fed to the LLM's transformer layers is therefore:
$\mathcal{H} \in \mathbb{R}^{\left( M + \sum_{i=1}^K N_i \right) \times d_{\text{text}}}.$
This strategy enables the LLM to process multiple, arbitrarily placed time series within a single, coherent context and capture complex inter-series and text-series dependencies. After the combination, the input embedding $\mathcal{H}$ is fed to the LLM to produce the final prediction $\mathcal{Y}$.

\subsection{Training Recipe}
\label{sec: training}
Our training process consists of two sequential stages: the first stage aligns time series data with the LLM to establish a foundational understanding of temporal-textual relationships, while the second stage refines the LLM’s reasoning capabilities to interpret and analyze these aligned representations. Throughout both stages, we keep the TSFM parameters frozen to preserve its pretrained temporal knowledge, while allowing the LLM’s parameters to remain trainable, ensuring adaptive learning without compromising the integrity of the encoded time series features.

\textbf{Stage 1: Pre-training for Language-Timeseries Alignment.} This stage aligns temporal data with textual information. We initially leverage synthesized data from~\citep{xie2024chatts}, which provides predefined templates to describe time series attributes. Although these template-based data offer accurate numerical information, their focus on specific time series patterns limits diversity, and the caption structure is monotonous. This lack of diversity can lead to overfitting to the templates, encouraging the model to learn shallow patterns and resulting in poor generalization~\citep{dong2025advances, choi2024voldoger}.
To alleviate this problem, we draw inspiration from captioning techniques in multimodal LLMs~\citep{chen2024sharegpt4v}. We synthesize comprehensive captions using advanced models (e.g., GPT-4.1) to enrich our alignment data. Specifically, we collect time series from two sources: \citep{merrill2024language}, which includes contextual information, and synthetic data from Chronos~\citep{ansari2024chronos}, which provides pure numerical time series.
\begin{figure*}[t]   
	
\centering

\includegraphics[width=1.0\textwidth]{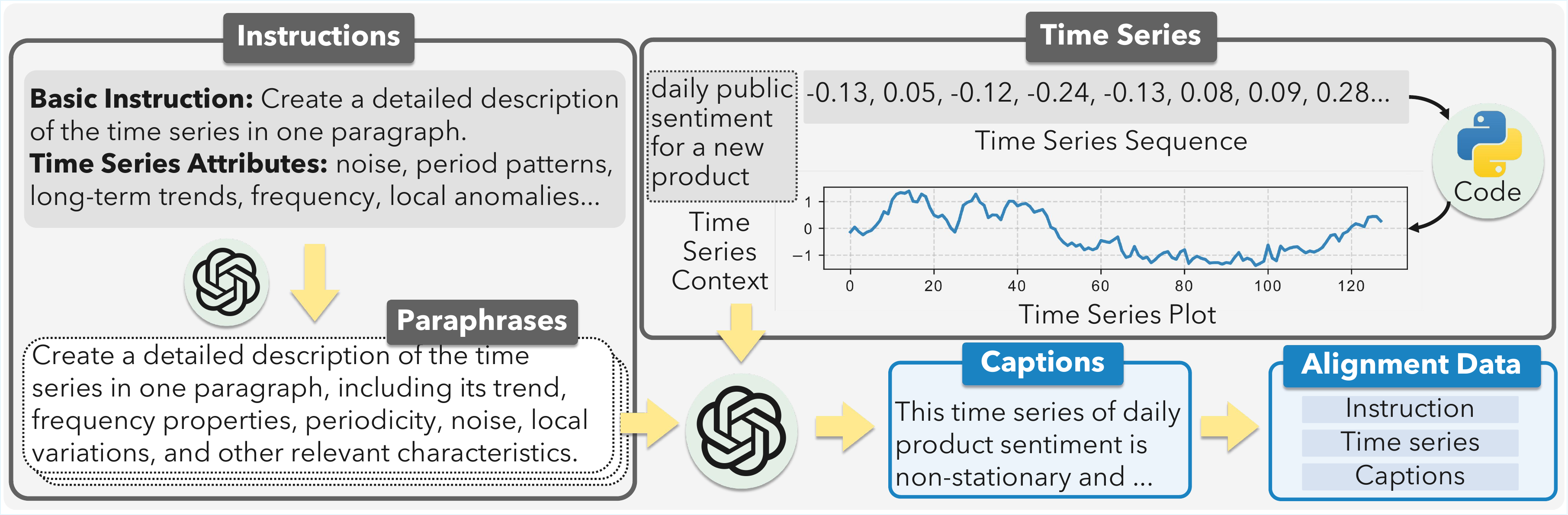
}

\caption{Workflow for our attribute-aware caption synthesis, designed to curate training data for alignment in stage 1. It enriches basic instructions with key attributes and generates diverse paraphrases, yielding high-fidelity captions to train \ours effectively.}

\label{fig:caption}
\vspace{-10pt}
\end{figure*}

\textbf{Attribute-aware Captioning.} Caption generation has been extensively investigated in visual domains~\citep{cheng2023beyond, cheng2025caparena, chen2024sharegpt4video}, playing a crucial role in multimodal alignment. However, time series captioning remains largely underexplored, presenting a significant impediment to achieving comprehensive alignment. To address this gap, we introduce a straightforward approach for generating scalable time series captions, as shown in Figure~\ref{fig:caption}.

Given a time series $\mathcal{T}$ with a temporal context $\C$, we begin by manually defining a generic captioning instruction, denoted as $\Ibase$. To facilitate enhanced comprehension by LLMs, we transform the time series into an image plot via Python code, $I_{TS} = \Phi(\mathcal{T})$. As evidenced in Table~\ref{tab:reasoning} (Section~\ref{sec: main results}), presenting the time series as an image to advanced LLMs (e.g., GPT-4.1) demonstrates a substantial advantage in understanding compared to providing it as a raw numerical series.

To enrich the generated captions, we augment $\Ibase$ with a fixed set of $G$ generic attribute categories of the time series, denoted as $\{a_1, a_2, \ldots, a_G\}$ (e.g., trend, frequency, periodicity, noise, local variations), yielding an augmented instruction $\Iprime = \Ibase \cup \{a_1, a_2, \ldots, a_G\}$. The instance-specific properties along them are inferred by the captioning LLM from each time-series visualization.
To further promote caption diversity, we use the LLM to paraphrase $\Iprime$ into $R$ distinct instructions, forming a candidate set of prompts $\mathcal{P} = \{\mathcal{I}''_{1}, \mathcal{I}''_{2}, \ldots, \mathcal{I}''_{R}\}$. For each time series $\mathcal{T}$, a prompt $\mathcal{I}''$ is uniformly sampled from this set, and the caption is generated conditioned on it and the time series visualization:
\begin{equation}
    \text{Caption} = \text{LLM}(\mathcal{I}'', I_{TS}, \C),
\end{equation}
where $\mathcal{I}'' \sim \mathcal{U}(\mathcal{P})$. The prompts are shown in Figure~\ref{fig:prompt-template-caption} in Appendix~\ref{sec:prompt}. We randomly sample 10K time series from each of two distinct sources: the Chronos synthetic dataset~\citep{ansari2024chronos}, which contains purely numerical time series, and a dataset of text-attributed time series from \citep{merrill2024language}, which provides contextual backgrounds. The construction of the data offers two benefits: 
(i) Pure time series data enables the model to build a foundational understanding of temporal patterns by focusing solely on the intrinsic characteristics of the data. 
(ii) Context-augmented time series enhances domain-specific comprehension by linking numerical trends to real-world scenarios.

\textbf{Stage 2: Instruction Fine-tuning for Time Series Reasoning.}
To elevate the model's capabilities from foundational understanding to complex reasoning, we employ an instruction fine-tuning stage based on the instruction tuning dataset~\citep{xie2024chatts}, which encompasses a wide range of Q\&As and instruction-following tasks. This training equips \ours with two critical abilities: the fidelity to adhere to complex instructions and structured response formats, and the capacity for nuanced, context-driven reasoning on time series-specific queries. 

\section{Experiments}
\textbf{Datasets.}
To assess the capabilities of \ours, we conduct comparative experiments against various baselines on benchmarks tailored for time series reasoning. Our evaluation incorporates TimeSeriesExam~\citep{cai2024timeseriesexam}, a comprehensive multiple-choice question answering dataset. TimeSeriesExam is specifically engineered to systematically evaluate a model's time series understanding and reasoning abilities across several key aspects: Pattern Recognition (PR), which addresses identifying trends, cycles, and stationarity; Noise Understanding (NU), focused on recognizing noise types such as white noise and random walks; Anomaly Detection (AD), for detecting unusual patterns; Similarity Analysis (SA), which involves comparing the shape and distribution of two time series; and Causality Analysis (CA), assessing the recognition of Granger Causality between time series. See Table~\ref{tab:reasoning_tasks} for more detailed descriptions. Furthermore, we evaluate on MTBench~\citep{chen2025mtbench}, a large-scale benchmark for evaluating time series reasoning in the real-world financial and weather domains, featuring questions that span both short-term (7-day) and long-term (14-day) temporal horizons, serving as a supplementary dataset since MTBench is a non-peer-reviewed preprint. Additionally, we evaluate an inductive reasoning task from~\citep{xie2024chatts} for open-ended reasoning ability.

\textbf{Baselines and Evaluation Metrics.}
We compare our method against three types of baselines: closed-source LLMs / VLMs, open-source LLMs / VLMs, and TSLLMs. Specifically, for closed-source models, we include GPT-4o, GPT-4.1~\citep{gpt4o, achiam2023gpt}, Claude-Sonnet-3.7~\citep{TheC3}, and DeepSeek-Chat~\citep{liu2024deepseek}. For open-source LLMs, we evaluate Llama-3.1-8B-Instruct~\citep{grattafiori2024llama}, Qwen2.5-7B-Instruct~\citep{yang2024qwen2}, GLM-4-9B-Chat~\citep{glm2024chatglm}, InternLM3-8B-Instruct~\citep{cai2024internlm2}, and Ministral-8B-Instruct~\citep{liu2026ministral}. Time series are transformed into textual sequences of numbers for LLMs. For open-source VLMs, we compare Qwen2.5-VL-7B~\citep{bai2025qwen2}, Phi-4-Multimodal-Instruct~\citep{abouelenin2025phi}, Llama3-LLaVA-Next-8B~\citep{li2024llavanext-strong}, InternVL3-8B~\citep{zhu2025internvl3}, and MiniCPM-V-2.6~\citep{yao2024minicpm}. Time series are transformed into plots via code for VLMs. For TSLLM models, we compare with ChatTime-7B~\citep{wang2025chattime}, ChatTS-14B~\citep{xie2024chatts}, and we use the official training data and code to fine-tune a 7B model for a fair comparison. 
As all benchmarks are multiple-choice Q\&As, we use accuracy as the evaluation metric. Reported accuracies are averaged over three inference runs of a single trained checkpoint using different random seeds. We further apply McNemar's test on the paired test-set predictions to assess the statistical significance of the improvements, and base our robustness claims on this test-set significance.

\label{sec: implementation}
\begin{wraptable}{r}{0.42\linewidth}
\vspace{-15pt}
\centering
\caption{Training details of \ours.}
\scalebox{0.95}{
\begin{tabular}{l|c|c}
\toprule
 & \textbf{Stage-1} & \textbf{Stage-2} \\
\midrule
Patch Size & 32 & 32 \\
Dataset & Captions & Instructions  \\
\#Samples & 120K & 30K \\ \midrule
TSFM & \multicolumn{2}{c}{TimesFM-1.0-200M} \\
LLM Backbone & \multicolumn{2}{c}{Qwen2.5-7B-Instruct} \\\midrule
Trainable Params & 7.3B & 7.3B  \\
Batch Size & 64 & 32  \\
Learning Rate & $1 \times 10^{-5}$ & $2 \times 10^{-5}$\\
Epoch & 1 & 2\\
\bottomrule
\end{tabular}
}
\label{tab:training-stages}
\end{wraptable}
\textbf{Implementation Details.} \ours uses the Qwen2.5-7B-Instruct as the LLM backbone across all the experiments with an embedding dimension of 3584, and uses the TimesFM-1.0-200M~\citep{das2024decoder} as our backbone TSFM with an embedding dimension of 1280. The TS-to-Text adapter is a single-layer MLP with a GELU activation, which projects the 1280-dimensional TSFM output features into the 3584-dimensional hidden space of the LLM. Parameters of the LLM are fine-tuned while those of the TSFM remain frozen during training. The detailed derivation of these time series embeddings from TimesFM can be found in Appendix~\ref{sec: timesfm}. We use $R=20$ different instructions to generate captions, and integrate $G=5$ different attributes into the instructions.
The training processes were conducted on 8 × L40s GPUs, where stage 1 and 2 consume about 20 and 5 hours, respectively. In stage 1, training data is composed of 100K template-based pairs and 20K LLM-generated caption pairs. In stage 2, we employ instruction tuning data from~\citep{xie2024chatts}. Comprehensive training parameters are further detailed in Table~\ref{tab:training-stages}. 

\subsection{Main Results}
\label{sec: main results}
Table~\ref{tab:reasoning} presents the performance of all models on the two benchmarks. The best results are bolded, and the second-best results are underlined.
Based on the results, we have the following key observations:

\textbf{(i) \ours achieves superior overall performance on all benchmarks among comparable-scale open-source models.}
Specifically, \ours surpasses the best-performing comparable-scale open-source LLM by 8.17 percentage points overall, the best comparable-scale open-source VLM by 5.82 points overall, and the best comparable-scale TSLLM by 4.11 points overall on the TimeSeriesExam benchmark. Compared to the backbone model, \ours improves on our backbone LLM performance by a substantial 8.17 points (a 17.5\% relative gain). \ours also outperforms the best comparable-scale open-source baseline on MTBench by 1.52--3.28 points. In addition, \ours also performs competitively against ChatTS-14B, which has a larger base model. We note that these claims are scoped to open-source models of comparable scale: proprietary models such as GPT-4.1 (vision) remain ahead on TimeSeriesExam overall accuracy, and we do not claim superiority over them. The notable improvement demonstrates the effectiveness of our model in various time series reasoning scenarios by introducing the temporal information of TSFM for the LLM.

(ii) \textbf{\ours shows consistent gains across time series reasoning subtasks.} \ours consistently leads across reasoning tasks, with notable improvements over the best comparable-scale open-source baseline in multiple subtasks like \textit{Pattern Recognition} (+3.33 points), \textit{Noise Understanding} (+7.28 points). It also gains 1.52--3.28 points in financial and weather reasoning. Such improvements stem from robust alignment and the enhanced ability to reason over numerical patterns within textual contexts.

Among all categories, Pattern Recognition shows the largest gap to GPT-4.1. We attribute this primarily to the general multi-step reasoning capacity gap between our 7B backbone and a frontier-scale model. This suggests that while \ours improves the model's time series understanding, its overall performance on tasks that demand both temporal understanding and complex multi-step reasoning remains bounded by the reasoning capacity of the base LLM. A promising direction for future work is to further strengthen this reasoning ability on top of the current model to narrow the gap to frontier models.

\begin{table*}[t]
\caption{
Performance on time series understanding and reasoning benchmarks. $\pm$ values are standard deviations over three inference runs of a single checkpoint. $^{*}$ marks the improvement of \ours over ChatTS-7B that are statistically significant under McNemar's test on the paired test-set predictions ($p<0.05$). We also include ChatTS-14B, which uses a larger base model. Bold and underline denote the best and second-best results among the comparable-scale open-source baselines.
}
\centering
\resizebox{\textwidth}{!}{
\begin{tabular}{l|cccccc | cccc}
\toprule
\multicolumn{1}{l|}{Model} & \multicolumn{6}{c|}{TimeSeriesExam~\citep{cai2024timeseriesexam}} & \multicolumn{4}{c}{MTBench~\citep{chen2025mtbench}} \\ \midrule
 & PR & NU & AD & SA & CA & OA & Finance (long) & Finance (short) & Weather (long) & Weather (short) \\
\midrule
\rowcolor{gray!20} \multicolumn{11}{c}{\textit{Proprietary models}} \\\midrule
DeepSeek-Chat & 65.23 & 55.17 & 52.71 & 63.71 & 42.86 & 59.89 &89.15 & 90.02 & 59.75 & 58.76\\
DeepSeek-R1 & 74.66 & 63.22 & 63.56 & 65.49 & 41.27 & 67.36 & 65.31 & 60.69 & 49.45 & 46.36\\
Claude-Sonnet-3.7 & 62.26 & 55.17 & 48.06 & 72.57 & 50.79 & 59.63 & 84.11 & 88.56 & 51.24 & 47.91\\
GPT-4o &  59.03 & 55.17 & 53.49 & 62.83 & 31.75 & 55.96 & 84.30 & 82.69 & 48.07 & 48.22 \\
GPT-4o (vision) & 67.12 & 62.07 & 62.79 & 64.60 & 26.98 & 62.12 & 84.11 & 80.65 & 46.43 & 48.53\\
GPT-4.1 (vision) & 69.81 & 68.97 & 68.22 & 75.22 & 41.27 & 67.89 & 93.41 & 91.45 & 56.04 & 55.35\\\midrule
\rowcolor{gray!20} \multicolumn{11}{c}{\textit{Open-source Large Language Models}} \\\midrule
Llama-3.1-8B-Instruct &  37.73 & 37.93 & 30.23 & 36.28 & 28.57 & 35.52 & 63.37 &35.52 & 40.25 & 40.00\\
Qwen2.5-7B-Instruct & 47.17 & 47.13 & 41.86 & 53.10 & \underline{41.27} & 46.66 & 87.98 & 89.41 & 57.14 &  58.44 \\
GLM-4-9B-chat &  41.78 & 39.08 & 37.21 & 47.79 & 38.09 & 41.28 & 71.31 & 77.19 & 50.27 & 50.85\\
InternLM3-8B-Instruct &  43.93 & 51.72 & 26.35 & 52.21 & 34.92 & 42.33 & 71.70 & 71.08 & 45.05 & 46.67\\
Ministral-8B-Instruct & 43.13 & 37.93 & 39.53 & 44.25 & 36.51 & 41.55 & 46.32 & 50.71 & 39.15 & 40.93\\\midrule
\rowcolor{gray!20} \multicolumn{11}{c}{\textit{Open-source Vision Language Models}} \\\midrule
Qwen2.5-VL-7B-Instruct & 25.34 & 32.18 & 19.38 & 42.48 & 12.70 & 26.61 & 81.98 & 86.35 & 52.06 & 46.82\\
Phi-4-Multimodal-Instruct  & 36.39 & 34.48 & 30.23	& 38.94 & 14.28 & 33.68 & 70.35 & 74.54 & 48.35 & 49.77 \\
Llama3-LLaVA-Next-8B & 31.27 & 35.63 & 29.46 & 30.09 & 38.09 & 31.85 & 52.14 & 51.50 & 47.53 & 47.29\\
InternVL3-8B & \underline{50.13} & \underline{52.87} & 43.41 & 54.87 & 38.09 & 49.01 & \underline{88.95} & \underline{90.00} & 53.71 & 54.88 \\
MiniCPM-V-2.6 & 29.11 & 39.08 & 27.13 & 51.33 & 31.75 & 33.42 & 81.78 & 83.09 & 48.63 & 45.12 \\\midrule
\rowcolor{gray!20} \multicolumn{11}{c}{\textit{Time Series Large Language Models}} \\\midrule
ChatTime-7B & 42.85 & 49.42 & 35.65 & 44.24 & 34.92 & 41.94 & 25.97 & 28.10 & 47.80 & 42.79\\
ChatTS -7B & \underline{50.13} & 50.57 & \underline{50.38} & \underline{61.95} & 34.92 & \underline{50.72} & 87.60 & 88.01 & \underline{58.92} & \underline{58.75} \\  
\textcolor{gray}{ChatTS -14B}  & \textcolor{gray}{59.30} & \textcolor{gray}{54.02} & \textcolor{gray}{51.16 }& \textcolor{gray}{62.83} & \textcolor{gray}{41.27} & \textcolor{gray}{56.36} & \textcolor{gray}{89.22} & \textcolor{gray}{91.22} & \textcolor{gray}{59.61} & \textcolor{gray}{59.22} \\\midrule
\ours -7B (ours)  & $\mathbf{53.46}^{*}_{\pm1.58}$ & $\mathbf{60.15}^{*}_{\pm3.51}$ & $\mathbf{53.23}^{*}_{\pm1.61}$ & $\mathbf{63.42}_{\pm1.02}$ & $\mathbf{43.39}^{*}_{\pm1.98}$ & $\mathbf{54.83}^{*}_{\pm0.98}$ & $\mathbf{92.00}^{*}_{\pm1.74}$ & $\mathbf{93.28}^{*}_{\pm1.28}$ & $\mathbf{60.44}^{*}_{\pm0.14}$ & $\mathbf{61.55}^{*}_{\pm0.31}$ \\
$\Delta$ Over Best ($\sim$7--9B) & +3.33 & +7.28 & +2.85 & +1.47 & +2.12 & +4.11 & +3.05 &  +3.28 & +1.52 & +2.80\\
\bottomrule
\end{tabular}
}
\label{tab:reasoning}
\vspace{-15pt}
\end{table*}

\subsection{Analysis of Data Scaling and Efficiency}
\begin{figure*}[t]   
	\centering
	\includegraphics[width=1.0\textwidth]{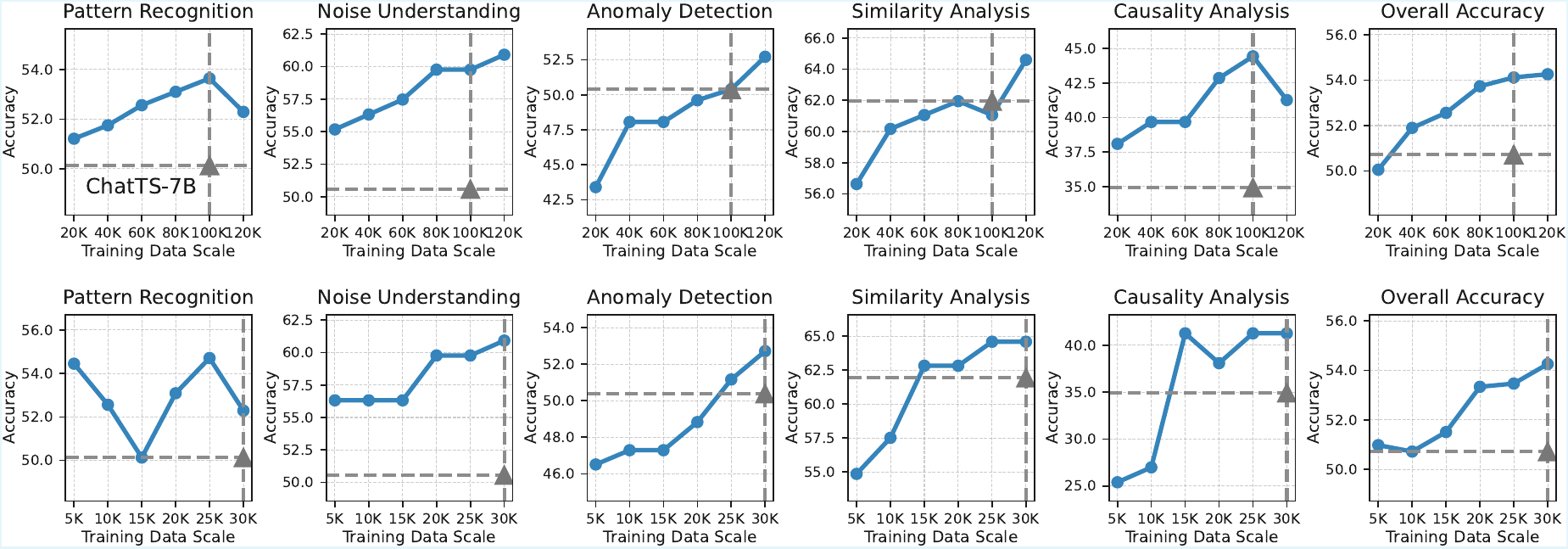}
	\caption{Data scaling and efficiency of \ours. The top (bottom) row illustrates how the performance of \ours varies when increasing the training data for alignment (instruction tuning). The columns correspond to sub-tasks in TimeSeriesExam. ChatTS-7B~\citep{xie2024chatts} is included for reference, denoted by the gray triangle.}
	\label{fig:data_scale}
	\vspace{-10pt}
\end{figure*}

Figure~\ref{fig:data_scale} presents our data scaling analysis on the TimeSeriesExam benchmark. \ours demonstrates remarkable data efficiency compared to the ChatTS-7B baseline. For the alignment stage, \ours achieves superior overall accuracy using just 60K samples, less than half the data required by the baseline. This efficiency is even more stark in the instruction tuning stage, where 10K samples suffice to outperform ChatTS-7B. This significant reduction in data dependency stems from our pre-trained TSFM and effective alignment, which equips the LLM with a robust temporal foundation. Consequently, \ours develops advanced reasoning capabilities with a substantially smaller amount of data, marking a key advantage for practical deployments where labeled data is scarce.

\subsection{Choices of Captioning Model for Alignment}
\begin{figure*}[t]
	\centering
	\includegraphics[width=1.0\textwidth]{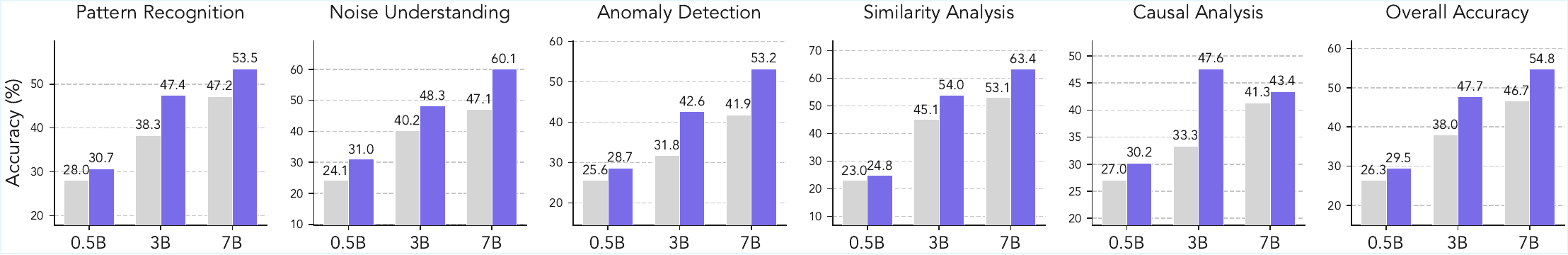}
	\includegraphics[width=1.0\textwidth]{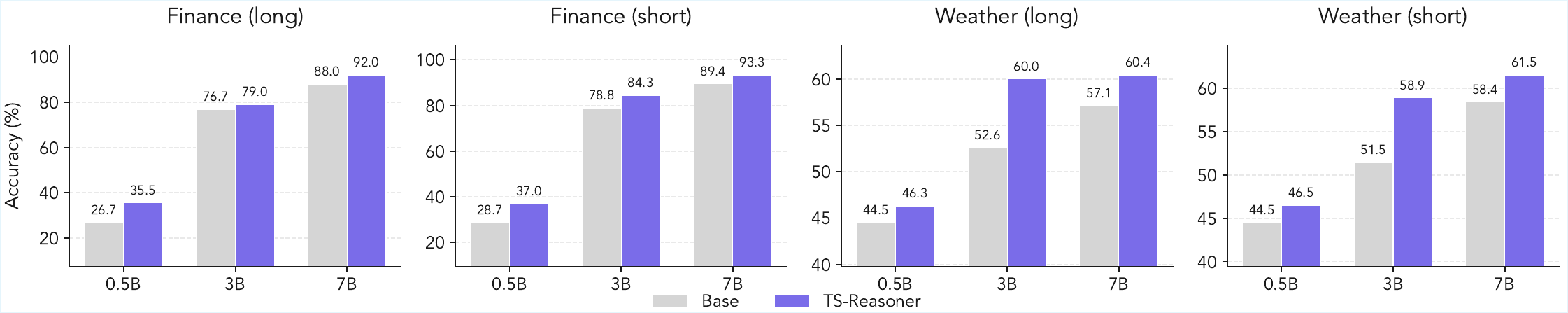}
	\caption{Performance of \ours and its associated LLM backbones (Qwen2.5 series). The top row and bottom row report the performance on TimeSeriesExam and MTBench, respectively.}
	\label{fig: model_size_mtbench}
	\vspace{-15pt}
\end{figure*}
The quality of the generated captions is a critical factor in the efficacy of our time-series-language alignment. To validate this, we trained \ours using three distinct sets of captioning data, each generated by a model with varying capabilities:
\begin{figure*}[t]   
    \centering
    \includegraphics[width=0.75\textwidth]{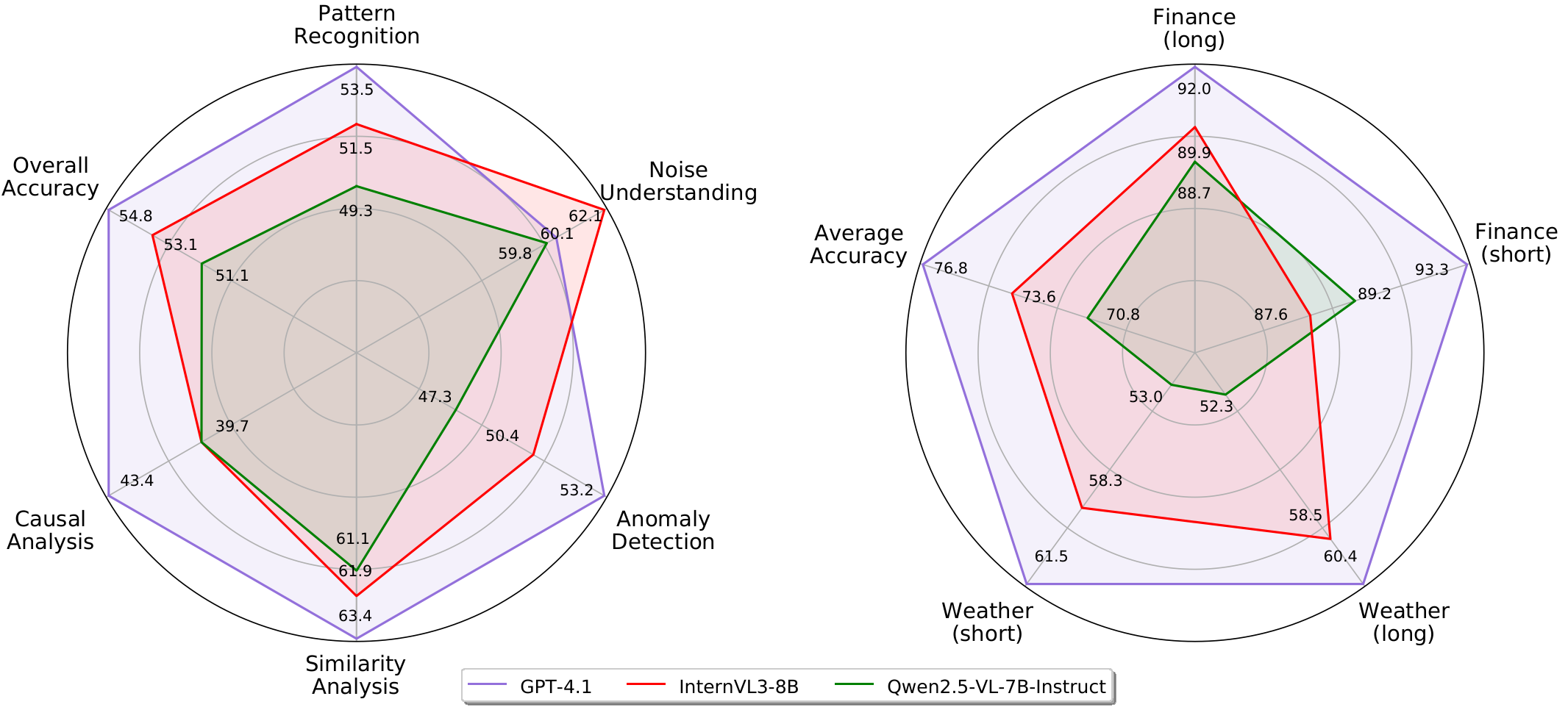}
    \caption{Comparison of multimodal LLMs used to generate time series captions for training \ours. \textbf{Left}: performance on TimeSeriesExam. \textbf{Right}: Performance on MTBench.}
    \label{fig:diff model time}
    \vspace{-20pt}
\end{figure*}
the state-of-the-art GPT-4.1, and two VLMs, InternVL3-8B and Qwen2.5-VL-7B-Instruct. As illustrated in Figure~\ref{fig:diff model time}, the results demonstrate that the performance of \ours is directly correlated with the fidelity of the captioning model. A distinct performance hierarchy emerges across both benchmarks: the model trained on GPT-4.1 captions consistently outperforms the one trained on InternVL3-8B captions, which in turn surpasses the one trained on Qwen2.5-VL-7B-Instruct captions. The higher performance gain from GPT-4.1 is attributed to its advanced capability in time series understanding. It is not surprising that the captions generated by InternVL3-8B achieve higher performance than Qwen2.5-VL-7B-Instruct, as its better time series understanding capability is shown in Table~\ref{tab:reasoning}. 

\subsection{Choices of TSFM and LLM in \ours}
\textbf{Different choices of TSFMs.}
To investigate the performance of \ours with different TSFMs, we replaced TimesFM (200M) with MOMENT-1-base (200M) and Chronos-base (200M), TSFMs of the same size, and re-evaluated its performance on the TimeSeriesExam benchmark. 
Results presented in Table~\ref{tab:diff_tsfm} reveal a substantial performance degradation when using MOMENT and Chronos, with overall accuracy falling from $54.83\%$ to $45.74\%$ and $53.21\%$, respectively. These results align with established forecasting benchmarks~\citep{shi2024time, mulayim2024time} where TimesFM demonstrates higher fidelity.
\begin{wraptable}{r}{0.6\columnwidth} 
    \centering
    \vspace{-6pt}
    \caption{Comparison of \ours using different TSFMs on the TimeSeriesExam benchmark.}
    \label{tab:diff_tsfm}
    \begin{tabular}{lcccccc}
        \toprule
        \textbf{Model} & \textbf{PR} & \textbf{NU} & \textbf{AD} & \textbf{SA} & \textbf{CA} & \textbf{OA} \\
        \midrule
        MOMENT & 46.90 & 47.13 & 41.86 & 54.87 & 28.57 & 45.74 \\
        Chronos & 51.75 & 59.77 & 51.93 & 63.71 & 36.51 & 53.21 \\
        TimesFM & $\textbf{53.46}$&$\textbf{60.15}$&$\textbf{53.23}$&$\textbf{63.42}$&$\textbf{43.39}$&$\textbf{54.83}$ \\
        \bottomrule
    \end{tabular}
    \vspace{-10pt}
\end{wraptable}
This suggests that TimesFM provides better time series representations, enabling \ours to understand and reason about time series better.

\textbf{Different choices of LLMs.}
To investigate the scalability and robustness of our approach with different LLM backbones, we evaluate \ours across three distinct sizes of the Qwen2.5-Instruct backbone: 0.5B, 3B, and 7B. The results, shown in Figure~\ref{fig: model_size_mtbench}, confirm that \ours is both highly effective and performs robustly. We observe a clear positive scaling law for both \ours and a baseline. More importantly, \ours maintains a consistent and significant lead across all models, with Overall Accuracy improvements of +3.15 points (29.49\% vs. 26.34\%), +9.70 points (47.71\% vs. 38.01\%), and +8.17 points (54.83\% vs. 46.66\%) for the 0.5B, 3B, and 7B models, respectively. This demonstrates that our approach performs robustly across different LLM backbones for complex time series reasoning. In addition, the performance gain from incorporating the TSFM is markedly smaller at 0.5B than at 3B and 7B, indicating that the TSFM's contribution grows with model scale. We speculate that this is because smaller models have insufficient capacity and limited base reasoning ability, which constrains how well they can align with and exploit the TSFM features. In contrast, larger models possess a higher-dimensional, semantically richer representation space and stronger context-integration and multi-step reasoning abilities, allowing them to more fully align, interpret, and exploit the injected temporal features, thereby converting the fine-grained temporal information into substantial downstream gains.

\subsection{Comparison between Textual and Visual Time Series for Captioning}
\begin{figure*}[t]   
	
\centering

\includegraphics[width=0.75\textwidth]{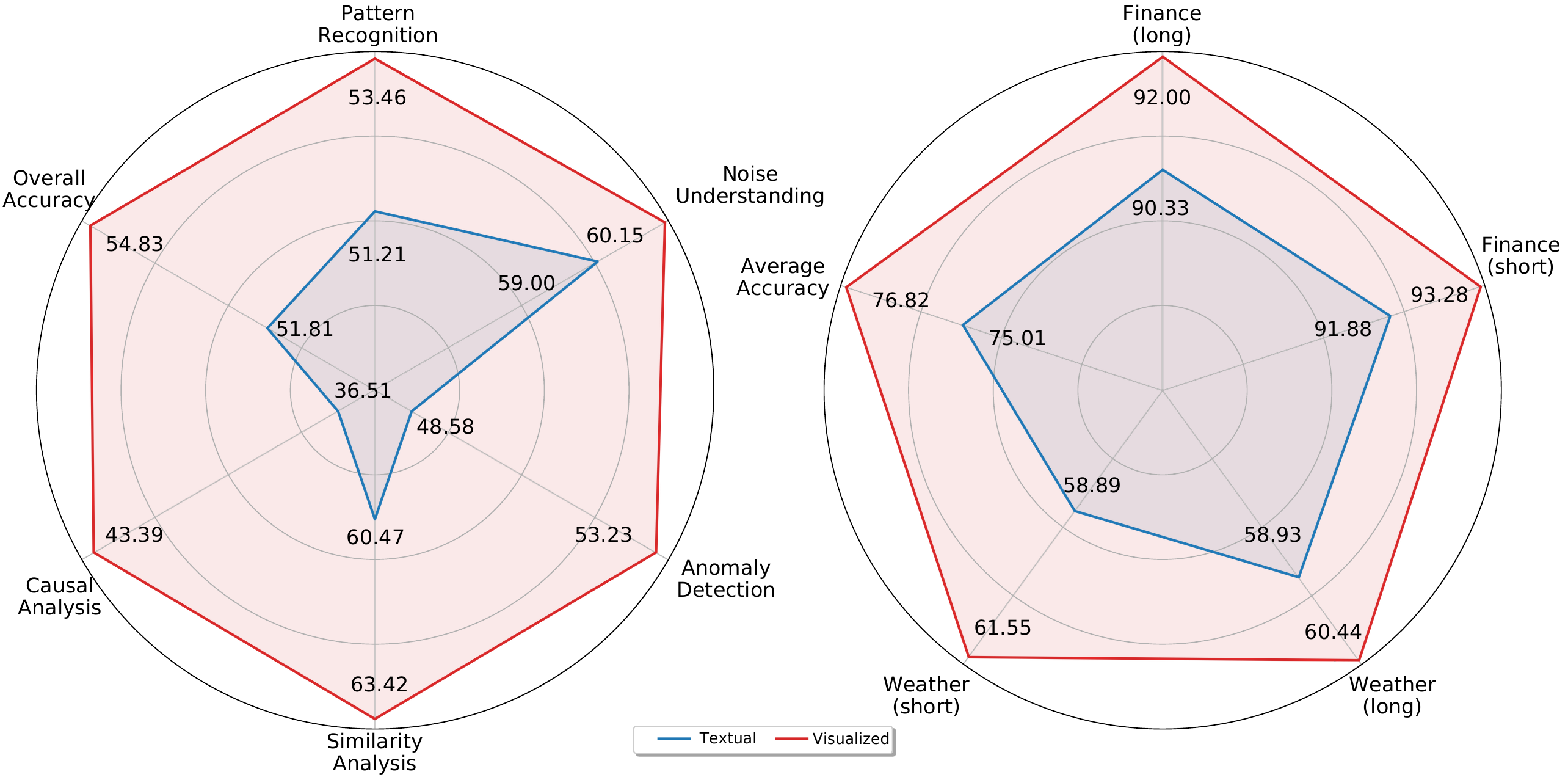}
\caption{Results of \ours on TimeSeriesExam (left) and MTBench (right) using textual and visualized time series for captioning as training data.}

\label{fig:text vision}
\vspace{-15pt}
\end{figure*}

To quantitatively assess the impact of different time series representations on captioning performance, we employed GPT-4.1 to synthesize training pairs based on either raw numerical sequences (textual) or visual plots (visual). As illustrated in Figure~\ref{fig:text vision}, \ours trained on visually-derived captions consistently outperforms its text-derived counterpart across most tasks. This indicates that the compact and holistic nature of visual plots enables LLMs to generate higher-fidelity captions, thereby facilitating more effective model alignment. To supplement these quantitative results, we provide a qualitative case analysis in Appendix~\ref{sec:case study}, which further demonstrates that visual time-series representations lead to a more accurate pattern recognition.

\subsection{Caption Diversity Analysis}
\label{sec: caption analysis}
A critical limitation of synthetic datasets is the risk of models learning spurious correlations from similar templates. To mitigate this, our attribute-aware generation process is designed to produce captions that are lexically diverse. To quantitatively validate the richness of our approach, we compare it against the template-based method. We evaluate both lexical diversity using the Measure of Textual Lexical Diversity (MTLD)~\citep{bestgen2024measuring} and Self-BLEU-4~\citep{zhu2018texygen} on a random sample of 1K captions from each dataset.
\begin{wraptable}{r}{0.5\columnwidth} 
    \centering
    \vspace{-10pt}
    \caption{Comparison of lexical diversity between template-based pairs and LLM-generated pairs.}
    \label{tab:diversity}
    \begin{tabular}{lcc}
        \toprule
        Metrics & MTLD $\uparrow$ & Self-BLEU-4 $\downarrow$ \\\midrule
        Template-based pairs & 42.95 & 0.82 \\
        LLM-generated pairs  & \textbf{133.30} & \textbf{0.45}\\
        \bottomrule
    \end{tabular}
    \vspace{-10pt}
\end{wraptable}
The results presented in Table~\ref{tab:diversity} show that our attribute-aware captions achieve an MTLD score of 133.30, a more than 3-fold increase over the template-based score of 42.95. Furthermore, the Self-BLEU-4 score is almost halved from 0.82 to 0.45. This substantial improvement in lexical diversity confirms that our method generates a significantly more expressive and diverse set of captions, crucial for training robust and generalizable models. We provide an example in Figure~\ref{fig:case study template} to compare the template-based caption and the LLM-generated caption.
To ensure comprehensive data coverage, we curated time series with context from a wide range of domains. The distribution of these domains is visualized in Figure~\ref{fig:statistics}.

\subsection{Open-ended Time Series Reasoning}
\label{sec: inductive}
\begin{wraptable}{r}{0.35\textwidth}
\vspace{-50pt}
\centering
\caption{Results on time series inductive reasoning.}
\vspace{-6pt}
\scalebox{0.95}{
\begin{tabular}{l|c}
\toprule
Model & Accuracy \\
\midrule
GPT-4o-mini & 33.30  \\
GPT-4o & 33.60\\
GPT-4o-mini (vision) & 32.30 \\
GPT-4o (vision) & 32.20\\
Qwen2.5-14B & 18.40\\
ChatTS-7B & 50.20\\
\textcolor{gray}{ChatTS-14B} & \textcolor{gray}{51.80}\\
\ours-7B & \textbf{54.70}\\
\bottomrule
\end{tabular}
}
\vspace{-18pt}
\label{tab:inductive}
\end{wraptable}

To evaluate \ours on open-ended time series reasoning tasks, we use the Inductive Reasoning dataset~\citep{xie2024chatts}, which requires the model to summarize the underlying physical principles reflected by uni/multivariate time series. Following~\citep{xie2024chatts}, we adopt RAGAS~\citep{es2024ragas}, an LLM-based keyword-matching evaluation framework. To verify its reliability, we randomly sample 100 cases for manual inspection and find 98\% agreement between the LLM evaluator and human judgments, confirming the trustworthiness of this evaluation protocol.

\noindent\textbf{Results.} As shown in Table~\ref{tab:inductive}, \ours-7B outperforms all baselines listed, exceeding GPT-4o and ChatTS-7B by 21.1 and 4.5 points, respectively. This substantial performance gap highlights the effectiveness of our approach in capturing complex temporal dynamics and performing reasoning on open-ended tasks. 

\subsection{Ablation Studies}
To further demonstrate the effectiveness of \ours, we conduct ablation studies to analyze the impact of individual components. Table~\ref{tab:ablation} summarizes our component-wise ablations from both training and model architecture perspectives:

(i) \textbf{Attribute-aware captioning is critical for robust language-timeseries alignment.} Removing the captioning data or stripping attributes from instructions degrades performance by up to 3.58 points on TimeSeriesExam and up to 3.00 points on average on MTBench. This confirms that fine-grained linguistic descriptions are essential for capturing nuanced temporal patterns. 

(ii) \textbf{Absence of any training stage significantly harms the performance.}
Removing Stage 1 (Alignment) primarily impacts MTBench's cross-modal reasoning tasks, while omitting Stage 2 (Instruction Tuning) causes a 29.02-point drop on TimeSeriesExam. This suggests that while alignment grounds the model, instruction tuning is essential for activating the ability to follow specific analytical commands. 

(iii) \textbf{Pretrained TSFM is beneficial for effective time series feature extraction.}
We remove the pretrained TSFM and repurpose the TS-to-Text adapter to directly project time series patches into the LLM's embedding space. As shown in Table~\ref{tab:ablation}, this modification leads to a performance decrease of 3.07 points on the TimeSeriesExam benchmark and 2.22 points on MTBench. We further replace time series embedding modules and use textual time series to train the base model with the same pipeline, which shows a 5.16-point decrease on the TimeSeriesExam benchmark and 2.60 points on MTBench. These results underscore the effectiveness of embedded time series and the importance of the TSFM as a beneficial temporal feature extractor. We additionally provide analysis on the scenarios where the TSFM is the most beneficial in Appendix~\ref{sec: benefit}. 

\begin{table*}[t!]
\centering
\caption{
Ablation study results of different components in \ours. $^{*}$ marks improvements that are statistically significant under McNemar's test ($p<0.05$) compared to the removal of TSFM.
}
\resizebox{\textwidth}{!}{
\begin{tabular}{l|cccccc | cccc}
\toprule
\multicolumn{1}{l|}{Model} & \multicolumn{6}{c|}{TimeSeriesExam~\citep{cai2024timeseriesexam}} & \multicolumn{4}{c}{MTBench~\citep{chen2025mtbench}} \\ \midrule
 & PR & NU & AD & SA & CA & OA & Finance (long) & Finance (short) & Weather (long) & Weather (short) \\
\midrule
\ours -7B  & $\textbf{53.46}^{*}$&$\textbf{60.15}^{*}$&$\textbf{53.23}^{*}$&$\textbf{63.42}$&$\textbf{43.39}^{*}$&$\textbf{54.83}^{*}$&$\textbf{92.00}^{*}$&$\textbf{93.28}^{*}$&$\textbf{60.44}^{*}$&$\textbf{61.55}^{*}$ \\\midrule
\rowcolor{gray!20} \multicolumn{11}{c}{\textit{Ablation on Training Data}} \\\midrule
- LLM-caption & 51.21 & 56.32 & 52.71 & 56.54 & 36.51 & 51.25 & 88.67 & 89.40 & 58.24 & 59.69\\
- Attributes & 52.02 & 57.47 & 48.83 & 62.83 & 39.68 & 52.69 & 89.71 & 89.20 & 57.28 & 59.07\\\midrule
\rowcolor{gray!20} \multicolumn{11}{c}{\textit{Ablation on Training Stages}} \\\midrule
- Stage 1 & 47.98 & 54.02 & 37.98 & 57.52 & 30.16 & 46.92  & 80.24 &  83.71 & 52.88 & 55.34 \\
- Stage 2 & 33.42 & 28.73 & 13.95 & 25.67 & 1.59 & 25.81 & 88.07 & 86.76 & 56.86 & 58.60\\\midrule
\rowcolor{gray!20} \multicolumn{11}{c}{\textit{Ablation on Model Architecture}} \\\midrule
- TSFM & 51.48 & 52.87 & 51.16 & 63.71 & 38.09 & 51.76 & 89.43 & 89.70 & 58.65 & 60.62 \\
- Freeze & 51.48 & 56.32 & 53.48 & 60.18 & 41.27 & 52.81 & 89.15 & 91.02 & 58.24 & 60.15 \\
TS-as-text & 	48.25 &55.17&42.64&53.98&50.45&49.67&88.95&90.43&58.10&59.38
\\

\bottomrule
\end{tabular}
}
\label{tab:ablation}
\vspace{-10pt}
\end{table*}
\section{Conclusion}
We introduce \ours, a framework that advances the ability of LLMs to understand and reason about time series by bridging with TSFM. To mitigate the intrinsic semantic gap, we further developed an attribute-aware captioning method that enriches time-series alignment data, fostering a more robust alignment. Extensive experiments demonstrate that \ours substantially outperforms a wide range of comparable-scale open-source LLMs, VLMs, and TSLLMs on time series understanding and reasoning benchmarks.

\noindent\textbf{Discussion.} In practice, the choice between \ours and proprietary models such as GPT-4.1 depends on deployment constraints. In domains such as finance and healthcare, data privacy regulations often prohibit sending time series to external cloud services, ruling out API-based proprietary models. As a fully open-source model that can be deployed locally, \ours is well-suited to these settings, while requiring substantially lower computational cost than large proprietary multimodal models. This makes \ours a practical choice when privacy, cost, or domain-specific adaptability are important considerations.




\bibliography{tmlr}

@article{cai2024timeseriesexam,
  title={TimeSeriesExam: A time series understanding exam},
  author={Cai, Yifu and Choudhry, Arjun and Goswami, Mononito and Dubrawski, Artur},
  journal={arXiv preprint arXiv:2410.14752},
  year={2024}
}

@inproceedings{du2022glam,
  title={Glam: Efficient scaling of language models with mixture-of-experts},
  author={Du, Nan and Huang, Yanping and Dai, Andrew M and Tong, Simon and Lepikhin, Dmitry and Xu, Yuanzhong and Krikun, Maxim and Zhou, Yanqi and Yu, Adams Wei and Firat, Orhan and others},
  booktitle={International conference on machine learning},
  pages={5547--5569},
  year={2022},
  organization={PMLR}
}

@article{yu2026tsrouter,
  title={TSRouter: Dynamic Modality-Model Selection for Time Series Reasoning},
  author={Yu, Fangxu and Feng, Tao and Min, Dehai and Cheng, Lu and Liu, Ge and Zhou, Tianyi},
  journal={arXiv preprint arXiv:2607.08940},
  year={2026}
}

@inproceedings{yu2026arrowgev,
  title={ArrowGEV: Grounding Events in Video via Learning the Arrow of Time},
  author={Yu, Fangxu and Lu, Ziyao and Niu, Liqiang and Meng, Fandong and Zhou, Jie},
  booktitle={Findings of the Association for Computational Linguistics: ACL 2026},
  pages={34657--34671},
  year={2026}
}

@article{yu2026reinforcement,
  title={Reinforcement Learning with Evolving Rubrics as Rewards for Audio Reasoning},
  author={Yu, Fangxu and Feng, Tao and Min, Dehai and Lin, Zinan and Xu, Weijia and Xu, Michael and Yu, Philip S and Liu, Ge and Zhou, Tianyi},
  journal={arXiv preprint arXiv:2608.02831},
  year={2026}
}

@article{yu2026weak,
  title={Weak-to-Strong On-Policy Distillation},
  author={Yu, Fangxu and Lin, Zinan and Liu, Xiaodong and Xu, Weijia and Xu, Michael and Zhou, Tianyi and Gao, Jianfeng},
  journal={arXiv preprint arXiv:2607.26246},
  year={2026}
}

@article{brown2020language,
  title={Language models are few-shot learners},
  author={Brown, Tom and Mann, Benjamin and Ryder, Nick and Subbiah, Melanie and Kaplan, Jared D and Dhariwal, Prafulla and Neelakantan, Arvind and Shyam, Pranav and Sastry, Girish and Askell, Amanda and others},
  journal={Advances in neural information processing systems},
  volume={33},
  pages={1877--1901},
  year={2020}
}

@article{liu2026ministral,
  title={Ministral 3},
  author={Liu, Alexander H and Khandelwal, Kartik and Subramanian, Sandeep and Jouault, Victor and Rastogi, Abhinav and Sad{\'e}, Adrien and Jeffares, Alan and Jiang, Albert and Cahill, Alexandre and Gavaudan, Alexandre and others},
  journal={arXiv preprint arXiv:2601.08584},
  year={2026}
}

@article{wei2021finetuned,
  title={Finetuned language models are zero-shot learners},
  author={Wei, Jason and Bosma, Maarten and Zhao, Vincent Y and Guu, Kelvin and Yu, Adams Wei and Lester, Brian and Du, Nan and Dai, Andrew M and Le, Quoc V},
  journal={arXiv preprint arXiv:2109.01652},
  year={2021}
}

@inproceedings{goswami2024moment,
  title={MOMENT: A Family of Open Time-series Foundation Models},
  author={Mononito Goswami and Konrad Szafer and Arjun Choudhry and Yifu Cai and Shuo Li and Artur Dubrawski},
  booktitle={International Conference on Machine Learning},
  year={2024}
}

@article{ho2025arcmemo,
  title={ArcMemo: Abstract Reasoning Composition with Lifelong LLM Memory},
  author={Ho, Matthew and Si, Chen and Feng, Zhaoxiang and Yu, Fangxu and Liu, Zhijian and Hu, Zhiting and Qin, Lianhui},
  journal={arXiv preprint arXiv:2509.04439},
  year={2025}
}

@article{cai2024internlm2,
  title={Internlm2 technical report},
  author={Cai, Zheng and Cao, Maosong and Chen, Haojiong and Chen, Kai and Chen, Keyu and Chen, Xin and Chen, Xun and Chen, Zehui and Chen, Zhi and Chu, Pei and others},
  journal={arXiv preprint arXiv:2403.17297},
  year={2024}
}

@article{liu2025sundial,
  title={Sundial: A family of highly capable time series foundation models},
  author={Liu, Yong and Qin, Guo and Shi, Zhiyuan and Chen, Zhi and Yang, Caiyin and Huang, Xiangdong and Wang, Jianmin and Long, Mingsheng},
  journal={arXiv preprint arXiv:2502.00816},
  year={2025}
}

@article{zhang2025timesbert,
  title={Timesbert: A bert-style foundation model for time series understanding},
  author={Zhang, Haoran and Liu, Yong and Qiu, Yunzhong and Liu, Haixuan and Pei, Zhongyi and Wang, Jianmin and Long, Mingsheng},
  journal={arXiv preprint arXiv:2502.21245},
  year={2025}
}

@article{achiam2023gpt,
  title={Gpt-4 technical report},
  author={Achiam, Josh and Adler, Steven and Agarwal, Sandhini and Ahmad, Lama and Akkaya, Ilge and Aleman, Florencia Leoni and Almeida, Diogo and Altenschmidt, Janko and Altman, Sam and Anadkat, Shyamal and others},
  journal={arXiv preprint arXiv:2303.08774},
  year={2023}
}

@inproceedings{jia2024gpt4mts,
  title={Gpt4mts: Prompt-based large language model for multimodal time-series forecasting},
  author={Jia, Furong and Wang, Kevin and Zheng, Yixiang and Cao, Defu and Liu, Yan},
  booktitle={Proceedings of the AAAI Conference on Artificial Intelligence},
  volume={38},
  number={21},
  pages={23343--23351},
  year={2024}
}

@article{chen2024sharegpt4video,
  title={Sharegpt4video: Improving video understanding and generation with better captions},
  author={Chen, Lin and Wei, Xilin and Li, Jinsong and Dong, Xiaoyi and Zhang, Pan and Zang, Yuhang and Chen, Zehui and Duan, Haodong and Tang, Zhenyu and Yuan, Li and others},
  journal={Advances in Neural Information Processing Systems},
  volume={37},
  pages={19472--19495},
  year={2024}
}

@inproceedings{cheng2023beyond,
  title={Beyond generic: Enhancing image captioning with real-world knowledge using vision-language pre-training model},
  author={Cheng, Kanzhi and Song, Wenpo and Ma, Zheng and Zhu, Wenhao and Zhu, Zixuan and Zhang, Jianbing},
  booktitle={Proceedings of the 31st ACM International Conference on Multimedia},
  pages={5038--5047},
  year={2023}
}

@article{cheng2025caparena,
  title={Caparena: Benchmarking and analyzing detailed image captioning in the llm era},
  author={Cheng, Kanzhi and Song, Wenpo and Fan, Jiaxin and Ma, Zheng and Sun, Qiushi and Xu, Fangzhi and Yan, Chenyang and Chen, Nuo and Zhang, Jianbing and Chen, Jiajun},
  journal={arXiv preprint arXiv:2503.12329},
  year={2025}
}

@article{yu2024flow,
  title={Flow of Reasoning: Training LLMs for Divergent Problem Solving with Minimal Examples},
  author={Yu, Fangxu and Jiang, Lai and Kang, Haoqiang and Hao, Shibo and Qin, Lianhui},
  journal={arXiv preprint arXiv:2406.05673},
  year={2024}
}

@article{wei2022chain,
  title={Chain-of-thought prompting elicits reasoning in large language models},
  author={Wei, Jason and Wang, Xuezhi and Schuurmans, Dale and Bosma, Maarten and Xia, Fei and Chi, Ed and Le, Quoc V and Zhou, Denny and others},
  journal={Advances in neural information processing systems},
  volume={35},
  pages={24824--24837},
  year={2022}
}

@article{fons2024evaluating,
  title={Evaluating Large Language Models on Time Series Feature Understanding: A Comprehensive Taxonomy and Benchmark},
  author={Fons, Elizabeth and Kaur, Rachneet and Palande, Soham and Zeng, Zhen and Balch, Tucker and Veloso, Manuela and Vyetrenko, Svitlana},
  journal={arXiv preprint arXiv:2404.16563},
  year={2024}
}

@article{hao2023reasoning,
  title={Reasoning with language model is planning with world model},
  author={Hao, Shibo and Gu, Yi and Ma, Haodi and Hong, Joshua Jiahua and Wang, Zhen and Wang, Daisy Zhe and Hu, Zhiting},
  journal={arXiv preprint arXiv:2305.14992},
  year={2023}
}

@article{liu2024lstprompt,
  title={Lstprompt: Large language models as zero-shot time series forecasters by long-short-term prompting},
  author={Liu, Haoxin and Zhao, Zhiyuan and Wang, Jindong and Kamarthi, Harshavardhan and Prakash, B Aditya},
  journal={arXiv preprint arXiv:2402.16132},
  year={2024}
}

@article{gruver2023large,
  title={Large language models are zero-shot time series forecasters},
  author={Gruver, Nate and Finzi, Marc and Qiu, Shikai and Wilson, Andrew G},
  journal={Advances in Neural Information Processing Systems},
  volume={36},
  pages={19622--19635},
  year={2023}
}

@article{merrill2024language,
  title={Language Models Still Struggle to Zero-shot Reason about Time Series},
  author={Merrill, Mike A and Tan, Mingtian and Gupta, Vinayak and Hartvigsen, Tom and Althoff, Tim},
  journal={arXiv preprint arXiv:2404.11757},
  year={2024}
}

@article{bai2025qwen2,
  title={Qwen2. 5-vl technical report},
  author={Bai, Shuai and Chen, Keqin and Liu, Xuejing and Wang, Jialin and Ge, Wenbin and Song, Sibo and Dang, Kai and Wang, Peng and Wang, Shijie and Tang, Jun and others},
  journal={arXiv preprint arXiv:2502.13923},
  year={2025}
}

@article{liu2024visual,
  title={Visual instruction tuning},
  author={Liu, Haotian and Li, Chunyuan and Wu, Qingyang and Lee, Yong Jae},
  journal={Advances in neural information processing systems},
  volume={36},
  year={2024}
}

@inproceedings{li2023blip,
  title={Blip-2: Bootstrapping language-image pre-training with frozen image encoders and large language models},
  author={Li, Junnan and Li, Dongxu and Savarese, Silvio and Hoi, Steven},
  booktitle={International conference on machine learning},
  pages={19730--19742},
  year={2023},
  organization={PMLR}
}

@article{zhang2025tempogpt,
  title={TempoGPT: Enhancing Temporal Reasoning via Quantizing Embedding},
  author={Zhang, Haochuan and Yang, Chunhua and Han, Jie and Qin, Liyang and Wang, Xiaoli},
  journal={arXiv preprint arXiv:2501.07335},
  year={2025}
}

@article{chow2024towards,
  title={Towards time series reasoning with llms},
  author={Chow, Winnie and Gardiner, Lauren and Hallgr{\'\i}msson, Haraldur T and Xu, Maxwell A and Ren, Shirley You},
  journal={arXiv preprint arXiv:2409.11376},
  year={2024}
}

@inproceedings{das2024decoder,
  title={A decoder-only foundation model for time-series forecasting},
  author={Das, Abhimanyu and Kong, Weihao and Sen, Rajat and Zhou, Yichen},
  booktitle={Forty-first International Conference on Machine Learning},
  year={2024}
}

@article{xie2024chatts,
  title={ChatTS: Aligning Time Series with LLMs via Synthetic Data for Enhanced Understanding and Reasoning},
  author={Xie, Zhe and Li, Zeyan and He, Xiao and Xu, Longlong and Wen, Xidao and Zhang, Tieying and Chen, Jianjun and Shi, Rui and Pei, Dan},
  journal={arXiv preprint arXiv:2412.03104},
  year={2024}
}

@inproceedings{wang2025chattime,
  title={Chattime: A unified multimodal time series foundation model bridging numerical and textual data},
  author={Wang, Chengsen and Qi, Qi and Wang, Jingyu and Sun, Haifeng and Zhuang, Zirui and Wu, Jinming and Zhang, Lei and Liao, Jianxin},
  booktitle={Proceedings of the AAAI Conference on Artificial Intelligence},
  volume={39},
  number={12},
  pages={12694--12702},
  year={2025}
}

@article{jin2023time,
  title={Time-llm: Time series forecasting by reprogramming large language models},
  author={Jin, Ming and Wang, Shiyu and Ma, Lintao and Chu, Zhixuan and Zhang, James Y and Shi, Xiaoming and Chen, Pin-Yu and Liang, Yuxuan and Li, Yuan-Fang and Pan, Shirui and others},
  journal={arXiv preprint arXiv:2310.01728},
  year={2023}
}

@article{chen2025mtbench,
  title={MTBench: A Multimodal Time Series Benchmark for Temporal Reasoning and Question Answering},
  author={Chen, Jialin and Feng, Aosong and Zhao, Ziyu and Garza, Juan and Nurbek, Gaukhar and Qin, Cheng and Maatouk, Ali and Tassiulas, Leandros and Gao, Yifeng and Ying, Rex},
  journal={arXiv preprint arXiv:2503.16858},
  year={2025}
}

@article{chen2024visionts,
  title={VisionTS: Visual Masked Autoencoders Are Free-Lunch Zero-Shot Time Series Forecasters},
  author={Chen, Mouxiang and Shen, Lefei and Li, Zhuo and Wang, Xiaoyun Joy and Sun, Jianling and Liu, Chenghao},
  journal={arXiv preprint arXiv:2408.17253},
  year={2024}
}

@article{cao2023tempo,
  title={Tempo: Prompt-based generative pre-trained transformer for time series forecasting},
  author={Cao, Defu and Jia, Furong and Arik, Sercan O and Pfister, Tomas and Zheng, Yixiang and Ye, Wen and Liu, Yan},
  journal={arXiv preprint arXiv:2310.04948},
  year={2023}
}

@misc{gpt4o,
    title = {OpenAI GPT-4o.},
    url = {https://openai.com/index/hello-gpt-4o/},
    year={2024}
}

@article{rb1990stl,
  title={STL: A seasonal-trend decomposition procedure based on loess},
  author={RB, CLEVELAND},
  journal={J Off Stat},
  volume={6},
  pages={3--73},
  year={1990}
}

@inproceedings{lai2024veclip,
  title={Veclip: Improving clip training via visual-enriched captions},
  author={Lai, Zhengfeng and Zhang, Haotian and Zhang, Bowen and Wu, Wentao and Bai, Haoping and Timofeev, Aleksei and Du, Xianzhi and Gan, Zhe and Shan, Jiulong and Chuah, Chen-Nee and others},
  booktitle={European Conference on Computer Vision},
  pages={111--127},
  year={2024},
  organization={Springer}
}

@article{vaswani2017attention,
  title={Attention is all you need},
  author={Vaswani, Ashish and Shazeer, Noam and Parmar, Niki and Uszkoreit, Jakob and Jones, Llion and Gomez, Aidan N and Kaiser, {\L}ukasz and Polosukhin, Illia},
  journal={Advances in neural information processing systems},
  volume={30},
  year={2017}
}

@inproceedings{li2022blip,
  title={Blip: Bootstrapping language-image pre-training for unified vision-language understanding and generation},
  author={Li, Junnan and Li, Dongxu and Xiong, Caiming and Hoi, Steven},
  booktitle={International conference on machine learning},
  pages={12888--12900},
  year={2022},
  organization={PMLR}
}

@article{wu2021autoformer,
  title={Autoformer: Decomposition transformers with auto-correlation for long-term series forecasting},
  author={Wu, Haixu and Xu, Jiehui and Wang, Jianmin and Long, Mingsheng},
  journal={Advances in Neural Information Processing Systems},
  volume={34},
  pages={22419--22430},
  year={2021}
}

@article{zhou2022fedformer,
  title={FEDformer: Frequency enhanced decomposed transformer for long-term series forecasting},
  author={Zhou, Tian and Ma, Ziqing and Wen, Qingsong and Wang, Xue and Sun, Liang and Jin, Rong},
  journal={arXiv preprint arXiv:2201.12740},
  year={2022}
}

@article{lim2021temporal,
  title={Temporal fusion transformers for interpretable multi-horizon time series forecasting},
  author={Lim, Bryan and Ar{\i}k, Sercan {\"O} and Loeff, Nicolas and Pfister, Tomas},
  journal={International Journal of Forecasting},
  volume={37},
  number={4},
  pages={1748--1764},
  year={2021},
  publisher={Elsevier}
}

@inproceedings{liu2021pyraformer,
  title={Pyraformer: Low-complexity pyramidal attention for long-range time series modeling and forecasting},
  author={Liu, Shizhan and Yu, Hang and Liao, Cong and Li, Jianguo and Lin, Weiyao and Liu, Alex X and Dustdar, Schahram},
  booktitle={International Conference on Learning Representations},
  year={2021}
}

@inproceedings{li2024transformer,
  title={Transformer-modulated diffusion models for probabilistic multivariate time series forecasting},
  author={Li, Yuxin and Chen, Wenchao and Hu, Xinyue and Chen, Bo and Zhou, Mingyuan and others},
  booktitle={The Twelfth International Conference on Learning Representations},
  year={2024}
}

@inproceedings{li2023prototype,
  title={Prototype-oriented unsupervised anomaly detection for multivariate time series},
  author={Li, Yuxin and Chen, Wenchao and Chen, Bo and Wang, Dongsheng and Tian, Long and Zhou, Mingyuan},
  booktitle={International Conference on Machine Learning},
  pages={19407--19424},
  year={2023},
  organization={PMLR}
}

@article{ltsm-bundle,
  title={Understanding Different Design Choices in Training Large Time Series Models},
  author={Chuang*, Yu-Neng and Li*, Songchen and Yuan*, Jiayi and Wang*, Guanchu and Lai*, Kwei-Herng and Yu, Leisheng and Ding, Sirui and Chang, Chia-Yuan and Tan, Qiaoyu and Zha, Daochen and Hu, Xia},
  journal={arXiv preprint arXiv:2406.14045},
  year={2024}
}

@article{sun2023test,
  title={Test: Text prototype aligned embedding to activate llm's ability for time series},
  author={Sun, Chenxi and Li, Hongyan and Li, Yaliang and Hong, Shenda},
  journal={arXiv preprint arXiv:2308.08241},
  year={2023}
}

@InProceedings{pmlr-v235-pan24c,
  title = 	 {$S^2${IP}-{LLM}: Semantic Space Informed Prompt Learning with {LLM} for Time Series Forecasting},
  author =       {Pan, Zijie and Jiang, Yushan and Garg, Sahil and Schneider, Anderson and Nevmyvaka, Yuriy and Song, Dongjin},
  booktitle = 	 {Proceedings of the 41st International Conference on Machine Learning},
  pages = 	 {39135--39153},
  year = 	 {2024},
  editor = 	 {Salakhutdinov, Ruslan and Kolter, Zico and Heller, Katherine and Weller, Adrian and Oliver, Nuria and Scarlett, Jonathan and Berkenkamp, Felix},
  volume = 	 {235},
  series = 	 {Proceedings of Machine Learning Research},
  month = 	 {21--27 Jul},
  publisher =    {PMLR},
  url = 	 {https://proceedings.mlr.press/v235/pan24c.html}
}

@article{prakarsha2022time,
  title={Time series signal forecasting using artificial neural networks: An application on ECG signal},
  author={Prakarsha, Kandukuri Ratna and Sharma, Gaurav},
  journal={Biomedical Signal Processing and Control},
  volume={76},
  pages={103705},
  year={2022},
  publisher={Elsevier}
}

@article{yang2024qwen2,
  title={Qwen2. 5 technical report},
  author={Yang, An and Yang, Baosong and Zhang, Beichen and Hui, Binyuan and Zheng, Bo and Yu, Bowen and Li, Chengyuan and Liu, Dayiheng and Huang, Fei and Wei, Haoran and others},
  journal={arXiv preprint arXiv:2412.15115},
  year={2024}
}

@article{liu2024moirai,
  title={Moirai-MoE: Empowering Time Series Foundation Models with Sparse Mixture of Experts},
  author={Liu, Xu and Liu, Juncheng and Woo, Gerald and Aksu, Taha and Liang, Yuxuan and Zimmermann, Roger and Liu, Chenghao and Savarese, Silvio and Xiong, Caiming and Sahoo, Doyen},
  journal={arXiv preprint arXiv:2410.10469},
  year={2024}
}

@article{nie2024survey,
  title={A survey of large language models for financial applications: Progress, prospects and challenges},
  author={Nie, Yuqi and Kong, Yaxuan and Dong, Xiaowen and Mulvey, John M and Poor, H Vincent and Wen, Qingsong and Zohren, Stefan},
  journal={arXiv preprint arXiv:2406.11903},
  year={2024}
}

@inproceedings{xu2023density,
  title={Density-aware temporal attentive step-wise diffusion model for medical time series imputation},
  author={Xu, Jingwen and Lyu, Fei and Yuen, Pong C},
  booktitle={Proceedings of the 32nd ACM International Conference on Information and Knowledge Management},
  pages={2836--2845},
  year={2023}
}

@article{dong2025advances,
  title={Advances in Multimodal Adaptation and Generalization: From Traditional Approaches to Foundation Models},
  author={Dong, Hao and Liu, Moru and Zhou, Kaiyang and Chatzi, Eleni and Kannala, Juho and Stachniss, Cyrill and Fink, Olga},
  journal={arXiv preprint arXiv:2501.18592},
  year={2025}
}

@article{choi2024voldoger,
  title={VolDoGer: LLM-assisted Datasets for Domain Generalization in Vision-Language Tasks},
  author={Choi, Juhwan and Kwon, Junehyoung and Yun, JungMin and Yu, Seunguk and Kim, YoungBin},
  journal={arXiv preprint arXiv:2407.19795},
  year={2024}
}

@article{zhong2025time,
  title={Time-VLM: Exploring Multimodal Vision-Language Models for Augmented Time Series Forecasting},
  author={Zhong, Siru and Ruan, Weilin and Jin, Ming and Li, Huan and Wen, Qingsong and Liang, Yuxuan},
  journal={arXiv preprint arXiv:2502.04395},
  year={2025}
}

@inproceedings{zerveas2021transformer,
  title={A transformer-based framework for multivariate time series representation learning},
  author={Zerveas, George and Jayaraman, Srideepika and Patel, Dhaval and Bhamidipaty, Anuradha and Eickhoff, Carsten},
  booktitle={Proceedings of the 27th ACM SIGKDD conference on knowledge discovery \& data mining},
  pages={2114--2124},
  year={2021}
}

@article{nie2022time,
  title={A time series is worth 64 words: Long-term forecasting with transformers},
  author={Nie, Yuqi and Nguyen, Nam H and Sinthong, Phanwadee and Kalagnanam, Jayant},
  journal={arXiv preprint arXiv:2211.14730},
  year={2022}
}

@article{liu2024timer,
  title={Timer: Generative pre-trained transformers are large time series models},
  author={Liu, Yong and Zhang, Haoran and Li, Chenyu and Huang, Xiangdong and Wang, Jianmin and Long, Mingsheng},
  journal={arXiv preprint arXiv:2402.02368},
  year={2024}
}

@article{ekambaram2024tiny,
  title={Tiny time mixers (ttms): Fast pre-trained models for enhanced zero/few-shot forecasting of multivariate time series},
  author={Ekambaram, Vijay and Jati, Arindam and Dayama, Pankaj and Mukherjee, Sumanta and Nguyen, Nam and Gifford, Wesley M and Reddy, Chandra and Kalagnanam, Jayant},
  journal={Advances in Neural Information Processing Systems},
  volume={37},
  pages={74147--74181},
  year={2024}
}

@article{grattafiori2024llama,
  title={The llama 3 herd of models},
  author={Grattafiori, Aaron and Dubey, Abhimanyu and Jauhri, Abhinav and Pandey, Abhinav and Kadian, Abhishek and Al-Dahle, Ahmad and Letman, Aiesha and Mathur, Akhil and Schelten, Alan and Vaughan, Alex and others},
  journal={arXiv preprint arXiv:2407.21783},
  year={2024}
}

@inproceedings{TheC3,
  title={The Claude 3 Model Family: Opus, Sonnet, Haiku},
  author={},
  url={https://api.semanticscholar.org/CorpusID:268232499}
}

@article{liu2024deepseek,
  title={Deepseek-v3 technical report},
  author={Liu, Aixin and Feng, Bei and Xue, Bing and Wang, Bingxuan and Wu, Bochao and Lu, Chengda and Zhao, Chenggang and Deng, Chengqi and Zhang, Chenyu and Ruan, Chong and others},
  journal={arXiv preprint arXiv:2412.19437},
  year={2024}
}

@inproceedings{chen2024sharegpt4v,
  title={Sharegpt4v: Improving large multi-modal models with better captions},
  author={Chen, Lin and Li, Jinsong and Dong, Xiaoyi and Zhang, Pan and He, Conghui and Wang, Jiaqi and Zhao, Feng and Lin, Dahua},
  booktitle={European Conference on Computer Vision},
  pages={370--387},
  year={2024},
  organization={Springer}
}

@article{yao2024minicpm,
  title={MiniCPM-V: A GPT-4V Level MLLM on Your Phone},
  author={Yao, Yuan and Yu, Tianyu and Zhang, Ao and Wang, Chongyi and Cui, Junbo and Zhu, Hongji and Cai, Tianchi and Li, Haoyu and Zhao, Weilin and He, Zhihui and others},
  journal={arXiv preprint arXiv:2408.01800},
  year={2024}
}

@article{zhu2025internvl3,
  title={Internvl3: Exploring advanced training and test-time recipes for open-source multimodal models},
  author={Zhu, Jinguo and Wang, Weiyun and Chen, Zhe and Liu, Zhaoyang and Ye, Shenglong and Gu, Lixin and Tian, Hao and Duan, Yuchen and Su, Weijie and Shao, Jie and others},
  journal={arXiv preprint arXiv:2504.10479},
  year={2025}
}

@misc{li2024llavanext-strong,
    title={LLaVA-NeXT: Stronger LLMs Supercharge Multimodal Capabilities in the Wild},
    url={https://llava-vl.github.io/blog/2024-05-10-llava-next-stronger-llms/},
    author={Li, Bo and Zhang, Kaichen and Zhang, Hao and Guo, Dong and Zhang, Renrui and Li, Feng and Zhang, Yuanhan and Liu, Ziwei and Li, Chunyuan},
    month={May},
    year={2024}
}

@article{bestgen2024measuring,
  title={Measuring lexical diversity in texts: The twofold length problem},
  author={Bestgen, Yves},
  journal={Language Learning},
  volume={74},
  number={3},
  pages={638--671},
  year={2024},
  publisher={Wiley Online Library}
}

@inproceedings{mulayim2024time,
  title={Are Time Series Foundation Models Ready to Revolutionize Predictive Building Analytics?},
  author={Mulayim, Ozan Baris and Quan, Pengrui and Han, Liying and Ouyang, Xiaomin and Hong, Dezhi and Berg{\'e}s, Mario and Srivastava, Mani},
  booktitle={Proceedings of the 11th ACM International Conference on Systems for Energy-Efficient Buildings, Cities, and Transportation},
  pages={169--173},
  year={2024}
}

@article{wang2025chronosteer,
  title={ChronoSteer: Bridging Large Language Model and Time Series Foundation Model via Synthetic Data},
  author={Wang, Chengsen and Qi, Qi and Rao, Zhongwen and Pan, Lujia and Wang, Jingyu and Liao, Jianxin},
  journal={arXiv preprint arXiv:2505.10083},
  year={2025}
}

@inproceedings{zhu2018texygen,
  title={Texygen: A benchmarking platform for text generation models},
  author={Zhu, Yaoming and Lu, Sidi and Zheng, Lei and Guo, Jiaxian and Zhang, Weinan and Wang, Jun and Yu, Yong},
  booktitle={The 41st international ACM SIGIR conference on research \& development in information retrieval},
  pages={1097--1100},
  year={2018}
}

@article{abouelenin2025phi,
  title={Phi-4-mini technical report: Compact yet powerful multimodal language models via mixture-of-loras},
  author={Abouelenin, Abdelrahman and Ashfaq, Atabak and Atkinson, Adam and Awadalla, Hany and Bach, Nguyen and Bao, Jianmin and Benhaim, Alon and Cai, Martin and Chaudhary, Vishrav and Chen, Congcong and others},
  journal={arXiv preprint arXiv:2503.01743},
  year={2025}
}

@article{glm2024chatglm,
  title={Chatglm: A family of large language models from glm-130b to glm-4 all tools},
  author={GLM, Team and Zeng, Aohan and Xu, Bin and Wang, Bowen and Zhang, Chenhui and Yin, Da and Zhang, Dan and Rojas, Diego and Feng, Guanyu and Zhao, Hanlin and others},
  journal={arXiv preprint arXiv:2406.12793},
  year={2024}
}

@article{shi2024time,
  title={Time-moe: Billion-scale time series foundation models with mixture of experts},
  author={Shi, Xiaoming and Wang, Shiyu and Nie, Yuqi and Li, Dianqi and Ye, Zhou and Wen, Qingsong and Jin, Ming},
  journal={arXiv preprint arXiv:2409.16040},
  year={2024}
}

@article{ansari2024chronos,
  title={Chronos: Learning the language of time series},
  author={Ansari, Abdul Fatir and Stella, Lorenzo and Turkmen, Caner and Zhang, Xiyuan and Mercado, Pedro and Shen, Huibin and Shchur, Oleksandr and Rangapuram, Syama Sundar and Arango, Sebastian Pineda and Kapoor, Shubham and others},
  journal={arXiv preprint arXiv:2403.07815},
  year={2024}
}

@article{qin2025cora,
  title={CoRA: Covariate-Aware Adaptation of Time Series Foundation Models},
  author={Qin, Guo and Chen, Zhi and Liu, Yong and Shi, Zhiyuan and Liu, Haixuan and Huang, Xiangdong and Wang, Jianmin and Long, Mingsheng},
  journal={arXiv preprint arXiv:2510.12681},
  year={2025}
}

@inproceedings{es2024ragas,
  title={Ragas: Automated evaluation of retrieval augmented generation},
  author={Es, Shahul and James, Jithin and Anke, Luis Espinosa and Schockaert, Steven},
  booktitle={Proceedings of the 18th conference of the european chapter of the association for computational linguistics: system demonstrations},
  pages={150--158},
  year={2024}
}

@article{yu2026tsrbench,
  title={TSRBench: A Comprehensive Multi-task Multi-modal Time Series Reasoning Benchmark for Generalist Models},
  author={Yu, Fangxu and Guo, Xingang and Yuan, Lingzhi and Kang, Haoqiang and Zhao, Hongyu and Qin, Lianhui and Huang, Furong and Hu, Bin and Zhou, Tianyi},
  journal={arXiv preprint arXiv:2601.18744},
  year={2026}
}
\bibliographystyle{tmlr}

\appendix

\section{TimesFM for time series embedding}
\label{sec: timesfm}
Given a time series $\mathcal{T} \in \mathbb{R}^{L}$, where $L$ is the length of the time series, we first normalize it to have a mean of zero and a variance of one. We then segment $\mathcal{T}$ into consecutive, non-overlapping patches of fixed length $P$, resulting in a total of $N = \left\lfloor L / P \right\rfloor$ patches. This yields a patched time series $\mathcal{T}_p \in \mathbb{R}^{N \times P}$.

Following the approach of~\citep{das2024decoder}, the $j$-th patch $\mathcal{T}_p^{j}$ is passed through a residual block to project it into the model dimension. This block is implemented as a two-layer MLP with a skip connection, processing each patch independently. The input token for the $j$-th patch is computed as:
\begin{equation}
\mathcal{E}_{p}^{j}= \text{InputResidualBlock}\left( \mathcal{T}_{p}^{j} \right) + \text{PE}_j,
\end{equation}
where $\text{PE}_j$ is the position encoding for the $j$-th patch, as defined in the original transformer~\citep{vaswani2017attention}. These encoded patch representations are then fed into an $M$-layer stacked Transformer to produce the final sequence of time series features:
\begin{equation}
    \mathcal{Z}_T = \text{StackedTransformer}([\mathcal{E}_p^{(0)}, \mathcal{E}_p^{(1)}, ..., \mathcal{E}_p^{(N)}]),
\end{equation}
where $\mathcal{Z}_T \in \mathbb{R}^{N \times d_{\text{ts}}}$ and $d_{\text{ts}}$ denotes the embedding dimension for each time series patch. Refer to more details of TimesFM in~\citep{das2024decoder}.

\begin{figure*}[h]   
	
\centering

\includegraphics[width=1.0\textwidth]{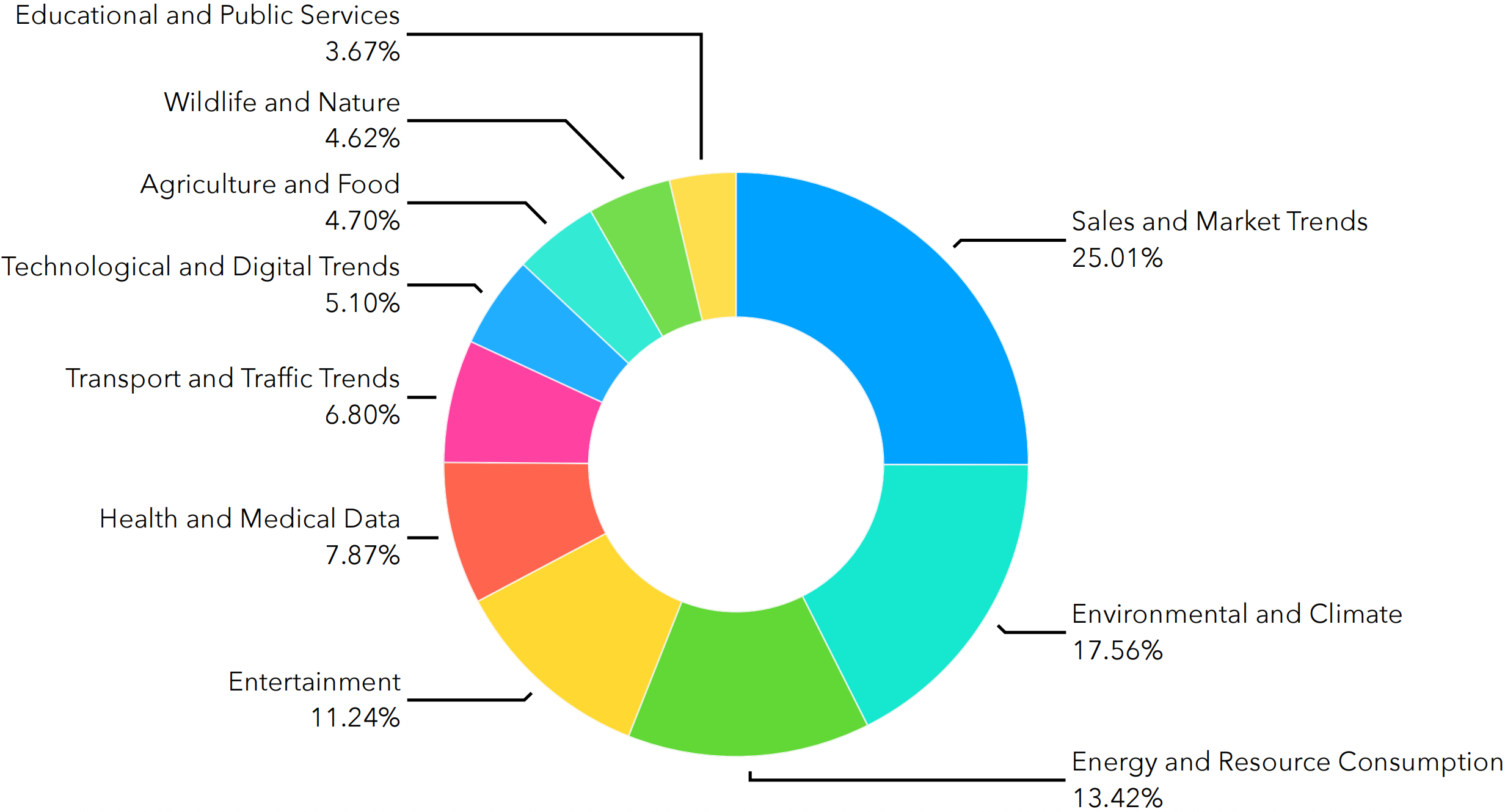}

\caption{Domain distribution of LLM-generated time series with context.}

\label{fig:statistics}
\end{figure*}

\section{Qualitative Analysis of Captioning}
\label{sec:case study}

\begin{figure*}[t]   
	
\centering

\includegraphics[width=1.0\textwidth]{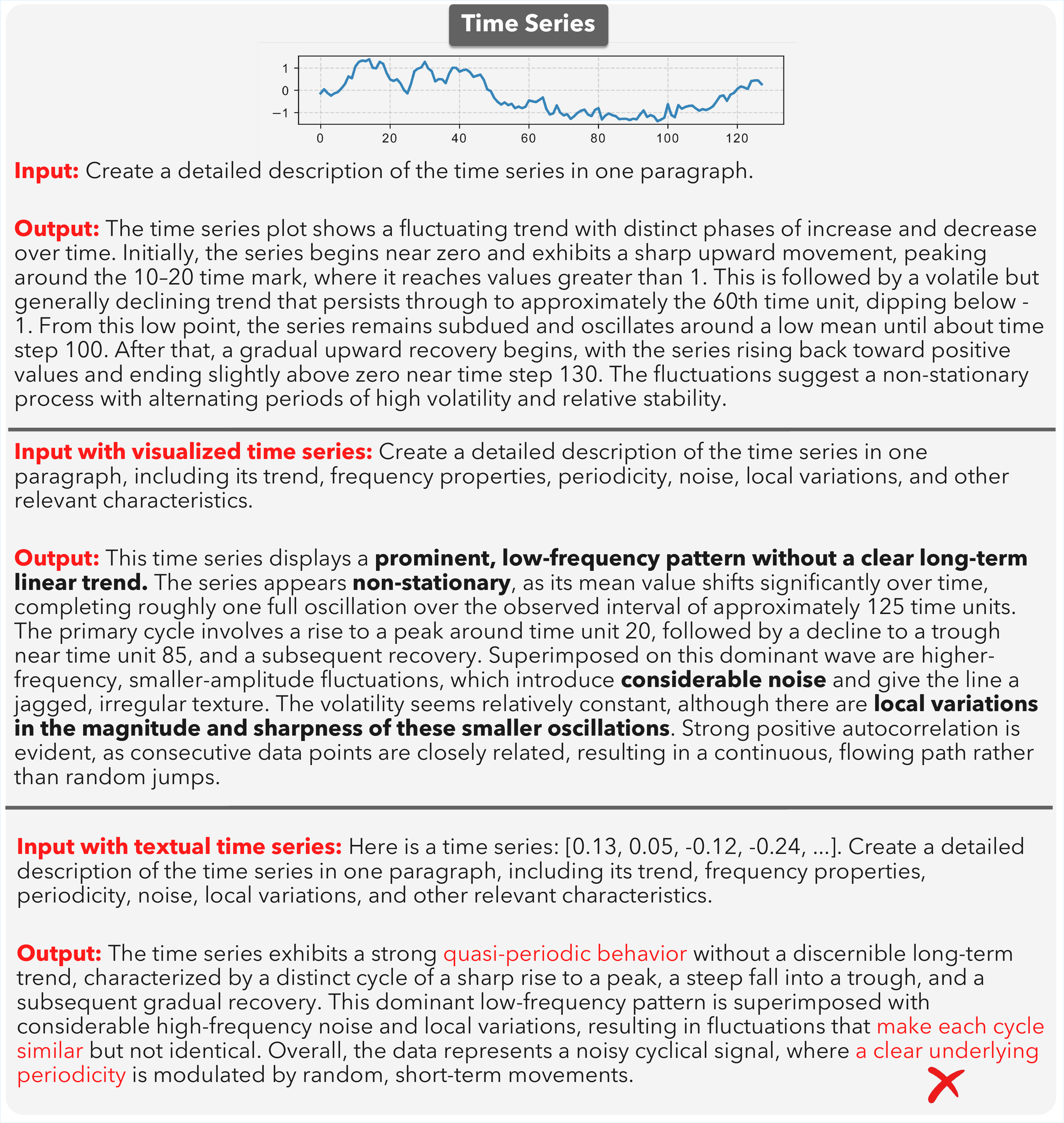}
\caption{Case study comparing time series captioning methods. Basic visual captioning (row 1) gives a chronological description. Our attribute-aware visual captioning (row 2) provides a richer description including key attributes. Textual time series captioning (row 3) identifies some attributes but fails to capture the periodicity pattern. }

\label{fig:case study}
	
\end{figure*}

To qualitatively evaluate the distinct advantages of our approach, we conduct a case study comparing three methods: (i) our proposed attribute-aware captioning, which leverages visual time series plots and explicit attribute guidance; (ii) a basic captioning baseline that operates on visual plots but lacks attribute guidance; and (3) LLM prompted with the raw textual (numerical) time series data. Our analysis, illustrated in Figure~\ref{fig:case study}, yields two key insights.

(i) \textbf{Attribute-Aware Captions Provide Semantically Richer Descriptions.} A primary limitation of basic captioning is its tendency to produce superficial, chronological narrations of the data. As shown in Figure~\ref{fig:case study}, the captioner describes the series' movements (e.g., "the value increases, then decreases sharply") but fails to extract deeper, underlying characteristics. While factually correct, this description omits properties crucial for a comprehensive understanding. In contrast, our attribute-aware captioning enriches this chronological account with critical semantic attributes. It not only captures the temporal dynamics but also identifies and articulates the series' overall trend, periodicity, and noise level. This multifaceted analysis provides a more holistic understanding of the time series, which is essential for \ours to conduct reasoning on downstream tasks. 

(ii) \textbf{Visual Representation is Crucial for Capturing Global Temporal Patterns.} When comparing our visually-grounded method to an LLM processing raw numerical data, a significant gap emerges in the ability to identify global patterns. The text-based LLM, while capable of discerning local features like high-frequency oscillations or noise within a limited window, consistently fails to recognize the overarching periodicity of the entire series. We hypothesize that this failure stems from the inherent inefficiency of representing long numerical sequences as text. The excessive length of input may distract the LLM, preventing it from observing the complete pattern. Conversely, a time series plot serves as a highly compressed, holistic representation. It enables the model to perceive the entire sequence as a single input, making global structures like periodicity visually salient and readily identifiable.

\begin{figure*}[t]   
	
\centering

\includegraphics[width=1.0\textwidth]{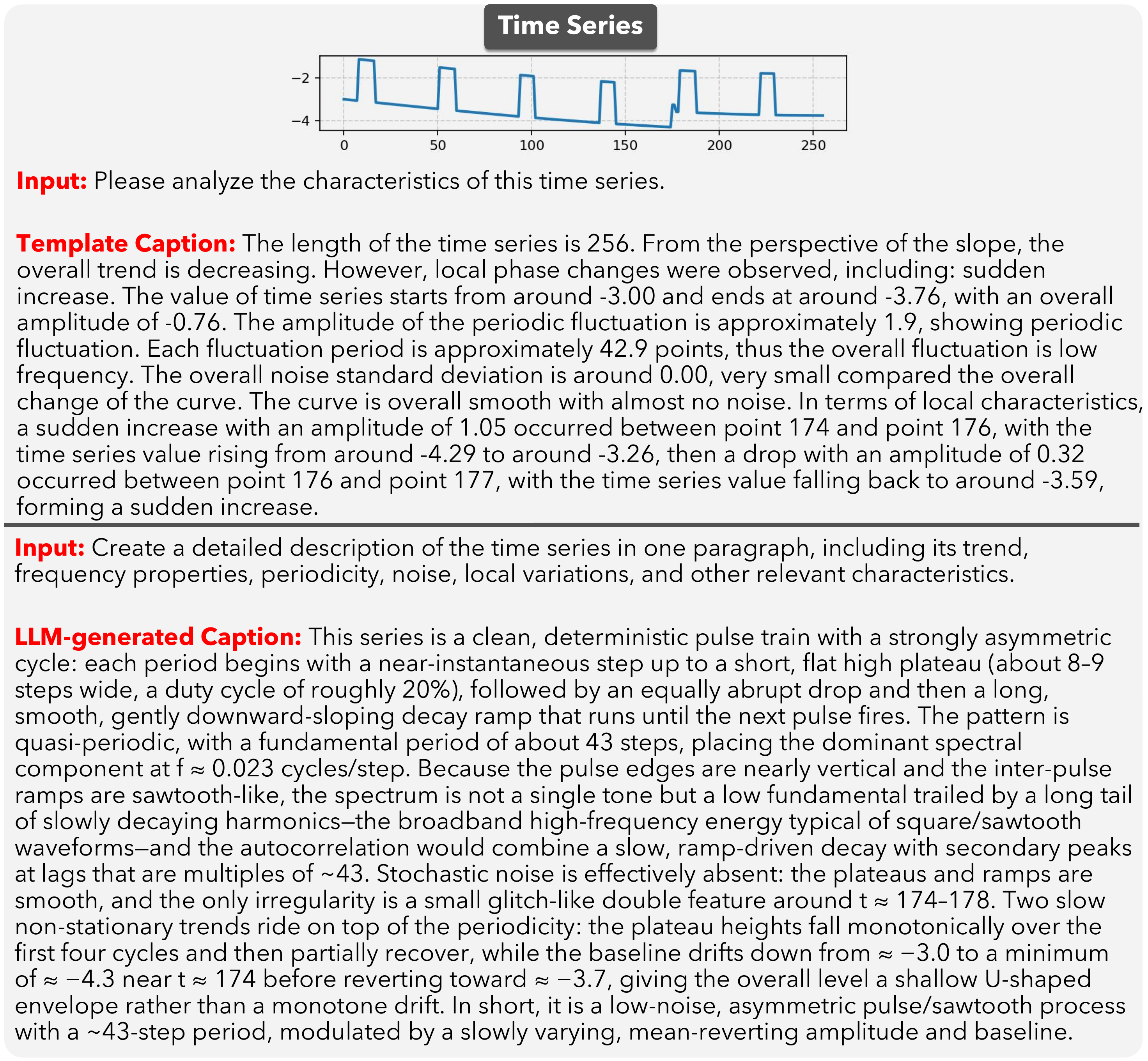}
\caption{Case study comparing time series captioning methods. Template-based captioning (row 1) fills a fixed set of predefined attributes with rigid, repetitive phrasing. Our LLM-generated captioning (row 2) offers greater language diversity and a more comprehensive, flexible description that is not restricted to predefined attributes.}

\label{fig:case study template}
	
\end{figure*}

\begin{table*}[t]
    \centering
    \caption{Example template questions for different reasoning tasks. Each subcategory covers a specific aspect of time series understanding, guiding the model to reason about comparative, anomalies, and causal relationships.}
    \label{tab:reasoning_tasks}
    \renewcommand{\arraystretch}{1.3}
    \begin{tabular}{l|c p{9cm}} 
        \toprule
        \textbf{Category} & \textbf{Subcategory} & \textbf{Example question} \\
        \midrule
        \multirow{10}{*}{Pattern Recognition} 
        & Trend & What is the most likely linear trend coefficient of the given time series? \\ \cmidrule{2-3}
        & Cyclic & The given time series has a sine wave pattern. \newline How does its amplitude change from the beginning to the end? \\ \cmidrule{2-3}
        & Stationarity & Is the given time series likely to be stationary after removing the cycle component? \\ \cmidrule{2-3}
        & Regime Switching & Based on the given time series, how many different regimes are there? \\ \cmidrule{2-3}
        & Statistical properties & Is the mean stable over time in the given time series? \\ \cmidrule{2-3}
        & Random processes & Does the following time series exhibit a mean reversion property? \\
        \midrule
        \multirow{5}{*}{Noise Understanding} 
        & White Noise & Is the given time series a white noise process? \\ \cmidrule{2-3}
        & Random Walk & Is the given time series likely to be a random walk process? \\ \cmidrule{2-3}
        & Signal / Noise Ratio & You are given two time series with the same underlying pattern but different noise levels. Which time series has a higher magnitude of noise? \\
        \midrule
        Anomaly Detection & & The following time series has two types of anomalies appearing at different time points. What are the likely types of these anomalies? \\
        \midrule
        \multirow{3}{*}{Similarity Analysis} 
        & Shape & Despite the noise, do the two given time series have similar patterns? \\ \cmidrule{2-3}
        & Distributional & You are given two time series, which are generated using a random walk. Are they likely to have the same variance? \\
        \midrule
        Causality Analysis & Granger Causality & Is there Granger causality between the two time series? \\
        \bottomrule
    \end{tabular}
\end{table*}
\newcolumntype{L}[1]{>{\raggedright\arraybackslash\hsize=#1\hsize}X}
\section{Comparison with Related LLM--Time-Series Architectures}
\label{sec:arch-comparison}

\begin{table}[t]
\centering
\footnotesize
\renewcommand{\arraystretch}{1.25}
\setlength{\tabcolsep}{4pt}
\begin{tabularx}{\textwidth}{@{} L{0.78} L{1.45} L{1.00} L{0.92} L{0.85} @{}}
\toprule
\textbf{Method} & \textbf{Architecture} & \textbf{Training Data} & \textbf{Frozen / Fine-tuned} & \textbf{Evaluation Tasks} \\
\midrule
ChronoSteer~\citep{wang2025chronosteer} &
LLM emits textual revision instructions to steer a TSFM's forecast (LLM$\rightarrow$TSFM) &
Synthetic instruction--series data &
LLM and TSFM frozen; alignment module trained &
Time series forecasting \\
\addlinespace[2pt]
TempoGPT~\citep{zhang2025tempogpt} &
VQ-VAE quantizes the series into tokens; the LLM vocabulary is expanded with a shared embedding layer &
Electrical time series simulation dataset &
VQ-VAE frozen; LLM fine-tuned &
Self-built electrical time series reasoning tasks \\
\addlinespace[2pt]
Time-VLM~\citep{zhong2025time} &
Time series encoded as images via a pretrained VLM, then fused for forecasting &
Public time series forecasting datasets &
VLM frozen; fusion network trained &
Time series forecasting \\
\addlinespace[2pt]
\textbf{\ours (ours)} &
Frozen pretrained TSFM features projected into the LLM via an adapter for reasoning (TSFM$\rightarrow$LLM) &
Cross-domain captions + instructions &
TSFM frozen; adapter + LLM fine-tuned &
Public reasoning benchmarks + open-ended tasks \\
\bottomrule
\end{tabularx}
\caption{Comparison of representative LLM--time-series architectures. Unlike
methods that steer a forecaster with text or quantize series into discrete
tokens, \ours aligns the \emph{continuous} embeddings of a frozen TSFM
with the LLM through a lightweight adapter, targeting reasoning rather than
forecasting alone.}
\label{tab:arch-comparison}
\end{table}

Recent work couples large language models (LLMs) with time series in several
distinct ways. Table~\ref{tab:arch-comparison} contrasts representative methods
along architecture, training data, which components are frozen versus
fine-tuned, and the evaluation tasks they target.

These methods differ in both the direction of knowledge transfer and the granularity of the time-series representation handed to the LLM.
ChronoSteer shows that LLM knowledge can be injected into a TSFM to strengthen
forecasting, transferring information in the LLM to the TSFM direction.
Conversely, \ours demonstrates the complementary direction: Integrating TSFM knowledge into the LLM improves the LLM's ability to understand time series. The two remaining baselines
differ from ours in representation granularity. TempoGPT tokenizes time series
into discrete tokens via a quantization codebook, whereas \ours aligns the
continuous embeddings of a pretrained TSFM with the LLM directly, avoiding the
information loss induced by quantization. Time-VLM renders the time series into
images and pair them with textual prompts to exploit a frozen VLM's
vision-language alignment for forecasting. In contrast, \ours grounds the
LLM in the continuous representations of a pretrained TSFM, showing that this is
an effective route to time-series understanding without relying on
visualization.

\section{Additional Experimental Results}

\begin{table}[t]
\centering
\caption{Effect of removing the TSFM on MTBench, broken down by domain
and average input length (number of time steps).
$\Delta_{\text{TSFM}}$ denotes the accuracy change of the ``--\,TSFM''
ablation relative to the full \ours (cf.\
Table~\ref{tab:ablation}). Within each domain, subsets with longer
input series incur larger drops. The ``short''/``long'' labels denote
the temporal horizon of the questions (7-day vs.\ 14-day).}
\label{tab:tsfm_length}
\begin{tabular}{llcc}
\toprule
Domain & Subset & Avg.\ time series length & $\Delta_{\text{TSFM}}$ \\
\midrule
\multirow{2}{*}{Finance} & short & 375 & $-3.58$ \\
                         & long  & 135 & $-2.57$ \\
\midrule
\multirow{2}{*}{Weather} & long  & 336 & $-1.79$ \\
                         & short & 168 & $-0.93$ \\
\bottomrule
\end{tabular}
\end{table}

\subsection{Scenarios Where TSFM is Most Beneficial}
\label{sec: benefit}
The ablation in Table~\ref{tab:ablation} shows that removing the TSFM
consistently degrades performance, though by a smaller margin than
removing instruction tuning. To clarify when the TSFM contributes
most, we analyze its effect along two axes.

\textit{(i) By input length.}
Table~\ref{tab:tsfm_length} reports the per-subset accuracy drop on
MTBench together with the average input length of each
subset.\footnote{In MTBench, the short/long split refers to the
question horizon (7-day vs.\ 14-day prediction), not the input length;
input lengths vary independently across subsets.}
Within each domain, the drop grows with input length: in finance,
removing the TSFM costs $3.58$ points at an average length of $375$
steps versus $2.57$ at $135$; in weather, $1.79$ at $336$ versus
$0.93$ at $168$. The TSFM is thus increasingly beneficial for longer
series, where its patch-level representations compactly summarize
long-range temporal structure that the LLM struggles to recover from
raw value sequences.

\textit{(ii) By modality reliance.}
Removing the TSFM reduces overall accuracy by $3.07$ points on
TimeSeriesExam but only $2.22$ points on average on MTBench
(Table~\ref{tab:ablation}). We attribute this gap to how strongly each
benchmark relies on the time-series modality: TimeSeriesExam questions
depend almost entirely on the time series itself, so weaker temporal
representations are directly penalized, whereas MTBench questions can
additionally draw on contextual news or weather reports that partially
compensate. The TSFM therefore contributes most when reasoning hinges
primarily on the time series, and more modestly when complementary
textual context is available.

\section{Prompts}
Below, we detail all the prompts we used. Figure~\ref{fig:prompt-template-caption} shows our captioning prompts. Figure~\ref{fig:prompt-timeseriesexam} shows the prompt for the TimeSeriesExam benchmark. Figure~\ref{fig:prompt-template-finance} and Figure~\ref{fig:prompt-template-weather} show the prompt for MTBench.
\label{sec:prompt}
\begin{figure*}
  \centering %
\begin{mybox}[Instructions for prompting LLMs to generate time series captions.]
\begin{obeylines}
1. Write a paragraph that analyzes the time series, covering its local behaviors, noise levels, periodic structures, overall trend, frequency content, and any other characteristics you consider important.

2. Create a detailed description of the time series in one paragraph, including its trend, frequency properties, periodicity, noise, local variations, and other relevant characteristics.

3. Provide a paragraph summarizing the time series characteristics such as noise, periodic patterns, long-term trends, frequency behavior, local anomalies, and any other significant features.

4.Compose a detailed caption describing the frequency characteristics, noise, trends, local variations, periodic structures, and any other meaningful patterns you observe in the time series.

5. Craft a one-paragraph summary of the time series, noting local fluctuations, periodic behavior, frequency features, trend, noise content, and any other insights you find important.

6. Generate a descriptive paragraph detailing the time series' key attributes, including frequency structure, noise patterns, trend direction, local features, periodic elements, and other notable aspects.

7. Give a thorough one-paragraph explanation of the time series, addressing periodicity, noise, frequency components, trend, local variations, and other relevant characteristics.

8. Write a narrative paragraph explaining the time series, focusing on noise, frequency characteristics, periodicity, localized structures, the overall trend, and other important features you identify.

9. Summarize the time series in a paragraph, describing its fluctuations, recurring patterns, noise levels, frequency-domain features, trend direction, and any additional traits you find significant.,
10. Develop a paragraph that captures the key features of the time series, such as frequency traits, trend, noise, periodic components, local behaviors, and other characteristics worth noting.
11. Provide a one-paragraph caption analyzing the time series data in terms of noise, trend, periodicity, local features, frequency-related behavior, and any additional characteristics of interest.
12. Create a rich paragraph description of the time series, including its trend, local anomalies, periodic activity, noise artifacts, spectral content, and other important descriptive elements.
13. Write a descriptive paragraph for the time series, highlighting frequency properties, trend behavior, periodic patterns, local structures, noise, and other characteristics you consider relevant.
14. Generate a compact yet thorough paragraph explaining the time series in terms of periodicity, trend movement, noise level, frequency details, local dynamics, and any other key aspects.
15. Construct a one-paragraph analysis of the time series by examining its local variations, noise, trend, periodic elements, frequency spectrum, and other notable features you deem important.
16. Write a summary paragraph that discusses the time series' periodic features, trend behavior, local patterns, noise levels, frequency domain signals, and other characteristics worth mentioning.,
17. Create a detailed one-paragraph commentary on the time series that outlines its noise characteristics, periodicity, frequency content, trends, localized behaviors, and other useful insights.
18. Prepare a paragraph-long description of the time series covering its trend, noise, frequency-related traits, local fluctuations, periodic structures, and any additional attributes of note.,
19. Offer a one-paragraph interpretation of the time series, highlighting its frequency features, periodic nature, local patterns, noise, trend line, and any other important characteristics you observe.
20. Compose a detailed summary in one paragraph focusing on the time series' periodic behavior, frequency spectrum, localized fluctuations, overall trend, noise, and other relevant descriptive elements.
\end{obeylines}
\end{mybox}
\vspace{-12pt}
\caption{The list of instructions for attributes-aware time series captioning.}
\label{fig:prompt-template-caption}

\end{figure*}

\begin{figure*}[t]
  \centering %
\begin{mybox}[Instructions for prompting LLMs for time series understanding and reasoning tasks in Timeseriesexam.]
\begin{obeylines}
You are a time series analysis expert. These are the time series data: $\langle\text{ts}_1\rangle\langle\text{ts}_1/\rangle \ldots \langle\text{ts}_k\rangle\langle\text{ts}_k/\rangle$. Please answer the question and provide the correct option letter.

Question: <Question>
Choices: <choices>
\end{obeylines}
\end{mybox}
\vspace{-12pt}
\caption{Prompt example for time series understanding and reasoning in Timeseriesexam.}
\label{fig:prompt-timeseriesexam}

\end{figure*}

\begin{figure*}[t]
  \centering %
\begin{mybox}[Instructions for prompting \ours for financial reasoning tasks in MTBench.]
\begin{obeylines}
You are an expert in finance and stock market analysis. Your task is to answer the question based on the given n-day historical stock price time series and a financial analysis published at the last timestamp of the time series.  The time series is: $\langle\text{ts}_1\rangle\langle\text{ts}_1/\rangle$.
<Context>
Question: <Question>
Choices: <Choices>
\end{obeylines}
\end{mybox}
\vspace{-12pt}
\caption{Prompt example for time series reasoning in Finance.}
\label{fig:prompt-template-finance}
\end{figure*}

\begin{figure*}[t]
  \centering %
\begin{mybox}[Instructions for prompting \ours for weather reasoning tasks in MTBench.]
\begin{obeylines}
You have an n-day temperature time series, and a weather event report published on the last day of the time series. The time series below is the 14-day temperature between <start time> to <end time>, and the time interval is 1 hour: $\langle\text{ts}_1\rangle\langle\text{ts}_1/\rangle$.  

The following events are reported:

<Context>
Question: <Question>
Choices: <Choices>
\end{obeylines}
\end{mybox}
\vspace{-12pt}
\caption{The prompt for time series reasoning in Weather.}
\label{fig:prompt-template-weather}
\end{figure*}

\section{Data Leakage Examination}
To guarantee evaluation integrity and prevent data leakage, particularly given LLM-synthesized data, we strictly prevent data leakage at both the source and semantic levels. We enforce source isolation by deriving training data exclusively from synthetic sources~\citep{ansari2024chronos} and public archives~\citep{merrill2024language} for alignment, and the instruction tuning dataset in~\citep{xie2024chatts}, explicitly verifying no numerical overlap with the MTBench and TimeSeriesExam benchmarks. Furthermore, we ensure narrative and template isolation; our synthesized training captions utilize attribute-aware prompts focusing on generic properties like trends and periodicity, which are structurally and semantically distinct from the domain-specific financial analyses and weather reports in MTBench and the systematic property evaluation questions in TimeSeriesExam.

In addition, we evaluate the potential overlap between the instruction-tuning data and the evaluation benchmarks by following~\citep{wei2021finetuned, brown2020language, du2022glam}, where an example from the evaluation benchmark is considered contaminated if any 13-gram in it also appears in the instruction-tuning data. Results in Table~\ref{tab:benchmark_rates} show that no example contains a 13-gram that appears in our Stage-2 instruction-tuning corpus. This 0\% contamination rate provides evidence that Stage 2 drives genuine reasoning capabilities rather than the memorization of benchmark-specific formats.

\begin{table*}[h]
\centering
\caption{Contamination ratio between instruction-tuning data and evaluation benchmark.}
\label{tab:benchmark_rates}
\resizebox{\textwidth}{!}{
\begin{tabular}{lccccc}
\toprule
\textbf{Benchmark} & TimeSeriesExam & MTBench QA-long (finance) & MTBench QA-long (weather) & MTBench QA-short (finance) & MTBench QA-short (weather) \\
\midrule
\textbf{Rate}      & 0.00\%         & 0.00\%                    & 0.00\%                    & 0.00\%                     & 0.00\%                     \\
\bottomrule
\end{tabular}
}
\end{table*}

\end{document}